\documentclass[11pt]{article}
\usepackage{booktabs}
\usepackage{multirow}
\usepackage[preprint]{acl}
\usepackage{times}
\usepackage{latexsym}
\usepackage[T1]{fontenc}
\usepackage[utf8]{inputenc}
\usepackage{microtype}
\usepackage{graphicx}
\usepackage{subcaption}
\usepackage[normalem]{ulem}
\newcommand{\revdel}[1]{}
\definecolor{experimentorange}{RGB}{230,120,20}

\hypersetup{
  colorlinks=true,
  linkcolor=red,
  citecolor=blue,
  urlcolor=blue
}

\usepackage[most]{tcolorbox}
\definecolor{takeawayyellow}{RGB}{255,248,220}
\newtcolorbox{takeaway}{
        colback=takeawayyellow,
        colframe=takeawayyellow,
        boxrule=0pt,
        arc=0pt,
        left=1mm,
        right=1mm,
        top=0.7mm,
        bottom=0.7mm,
        before skip=4pt,
        after skip=4pt
}

\usepackage{amsmath,amsfonts,bm}

\def\eqref#1{equation~\ref{#1}}
\def\1{\bm{1}}

\DeclareMathAlphabet{\mathsfit}{\encodingdefault}{\sfdefault}{m}{sl}
\SetMathAlphabet{\mathsfit}{bold}{\encodingdefault}{\sfdefault}{bx}{n}

\title{LatentRemedy: Efficient Remedy For Non-Prefix KV Cache Reuse in Multi-Agent Systems}

\author{
Zhiwei Chen\textsuperscript{1}, Zhongtao Rao\textsuperscript{1,2}, Waikit Xiu\textsuperscript{1,3}, \\
\textbf{Yuxiang Zhang\textsuperscript{1}, Shuting Cui\textsuperscript{1}, Jian Chen\textsuperscript{3}, Tianxiang Zhao\textsuperscript{1}\thanks{Corresponding author.}} \\
\textsuperscript{1}The Hong Kong University of Science and Technology (Guangzhou) \\
\textsuperscript{2}Microsoft \\
\textsuperscript{3}The University of Hong Kong
}

\begin{document}

% Keep corresponding-author footnote links black without changing body links.
\hypersetup{linkcolor=black}
\maketitle
\hypersetup{linkcolor=red}

\begin{abstract}
% mas中重复文本提供kv cache reuse的机会
Non-prefix KV cache reuse is a promising approach for improving inference efficiency in multi-agent systems as agents share the same context blocks. 
% 当agent冲突时reuse会影响性能
However, when agents have conflicting tasks, directly reusing non-prefix KV caches across agents can severely degrade model performance, causing up to a 90\% drop.
% 现有的 cache repair 方法主要修复 KV representation，但在 conflicting tasks 下效果有限
Existing remedies, including selective recomputation and offset-based calibration, recover a limited fraction of this loss.
% 
% 核心 insight：问题不仅在于 KV representation 不准确，还在于 target agent 无法正确利用 reused context 完成自己的任务
We argue that the problem lies not only in contaminated KV representations but also in the agent's difficulty interpreting reused context under its own instructions.
% 
% 提出 LatentRemedy：使用少量包含 target instruction 信息的 latent tokens，在不重新计算 shared context 的情况下帮助模型完成目标任务
To this end, we introduce LatentRemedy, which uses a few task-conditioned latent tokens to help the target agent leverage reused context without recomputing the shared block.
% 简述三个设计：从 target prefix 生成 tokens、放在 reused block 后交互、利用 distillation 指导后续生成
Specifically, we use an MLP to map the last hidden state of the target prefix into a small number of latent tokens, which are placed after the reused KV block to interact with the shared context. To guide subsequent predictions toward the target task, we train the MLP through full-prefill distillation.
% 总结 accuracy 和 TTFT 的实验优势
Experiments show that LatentRemedy consistently outperforms existing reuse baselines on conflicting tasks while maintaining TTFT close to direct reuse as cache size grows. Code and data are available at \url{https://github.com/chenzhwsysu57/kvreuse}.

\end{abstract}

\section{Introduction}
\begin{figure}[t]
\centering
\includegraphics[width=\linewidth]{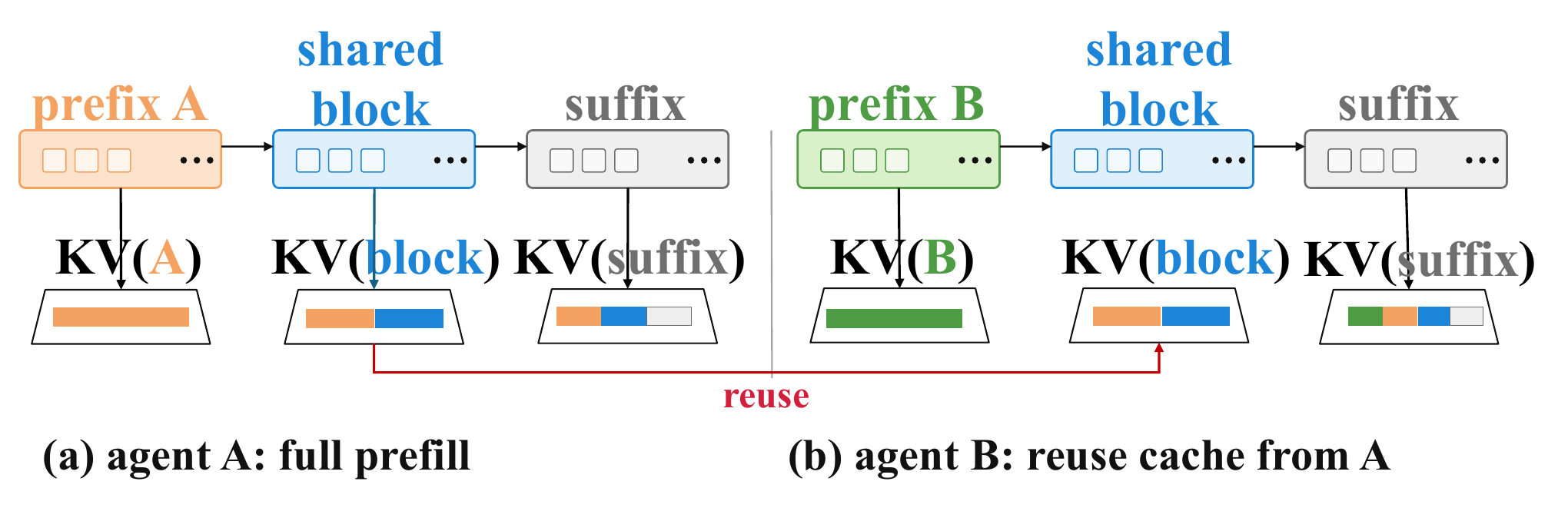}
\caption{Illustration of reusing KV-cache of the same block from task A to task B. }
\label{fig:illustrate-kv-reuse}
\end{figure}

KV-cache reuse is an effective way to reduce redundant prefill computation \cite{yao2025cacheblend,kvcomm2025,geng2026relaycaching,ramesh2025communicating} in multi-agent systems where agents share the same context blocks. However, existing work largely focuses on cooperative agents with aligned objectives, leaving agents with \textbf{conflicting objectives} rarely explored. As illustrated in Figure~\ref{fig:illustrate-kv-reuse}, agent A and agent B may process the same context block under different or even conflicting task instructions. Since the shared block's KV cache is computed under agent A's instruction, reusing it for agent B may interfere with how agent B uses the shared context to perform its own task.

We find that existing position-independent cache repair methods, which rely on selective recomputation or offset-based calibration, recover only a limited fraction of the performance loss under conflicting tasks. Our analysis suggests that previous agent's task in the reused KV cache can interfere with how the current agent interprets the shared context. Unlike full prefill, where the shared context is processed under the agent's instruction, cross-agent KV reuse introduces representations computed under a different task. This suggests that an effective remedy should allow the instruction to interact with the reused context, while ensuring that this interaction guides subsequent generation towards current objective. Our preliminary study further shows that repeating the target instruction after the reused cache can substantially alleviate task interference.

To this end, we introduce \textbf{LatentRemedy}, which addresses this problem with a small number of latent tokens. To encode the prefix instruction, we use an MLP to map the last hidden state of the prefix into these latent tokens. We place them after the reused block, allowing them to attend to both the prefix and the reused context. To guide consecutive reasoning towards the prefix instruction, we train the MLP through full-prefill distillation. Extensive experiments show that LatentRemedy substantially outperforms existing KV cache reuse methods in conflicting-task scenarios, while maintaining time-to-first-token (TTFT) latency close to direct reuse.
In a nutshell, our contributions are threefold:
\begin{itemize}
    \item We identify \emph{conflicting KV-cache reuse} scenarios in multi-agent systems, where existing cache repair methods struggle to recover the performance loss.
    \item We reveal that cross-agent KV cache reuse disrupts target-spefic contextualization, highlighting the needs of prefix re-conditioning when reusing non-prefix KV cache.
    \item We introduce LatentRemedy, a simple latent remedy that enables interaction with reused context to guide subsequent generation without recomputing the shared block. Experiments demonstrate its advantages in accuracy and TTFT across various tasks.
\end{itemize}

\section{Related Work}

\textbf{Non-prefix KV-cache reuse.} Methods such as CacheBlend, KVCOMM, RelayCaching, and EPIC repair cached states by selective recomputation or offset recalibration~\cite{yao2025cacheblend,kvcomm2025,geng2026relaycaching,hu2025epic}, primarily address cache reuse in cooperative scenarios, whereas applications in non-cooperative scenarios have rarely been explored.  

\noindent \textbf{Soft prompt tuning.} 
By optimizing a small set of continuous embeddings prepended to the input, such methods offer an efficient way of full-model fine-tuning~\cite{lester-etal-2021-power, li-liang-2021-prefix, liu-etal-2022-p-tuning}. Unlike these methods that learn fixed task-level prompts for general adaptation, our remedy seeks to generate request-specific latent tokens from the target prefix to guide task-specific generation.

\section{Preliminary Study}
\label{sec:preliminary-study}
% 上下子图
% figure/no_reasoning_argkp_deal_job_interview.pdf
% figure/reasoning_argkp_deal_job_interview.pdf
\begin{figure}[t]
    \centering
    \begin{subfigure}{1\linewidth}
        \centering
        \includegraphics[width=\linewidth]{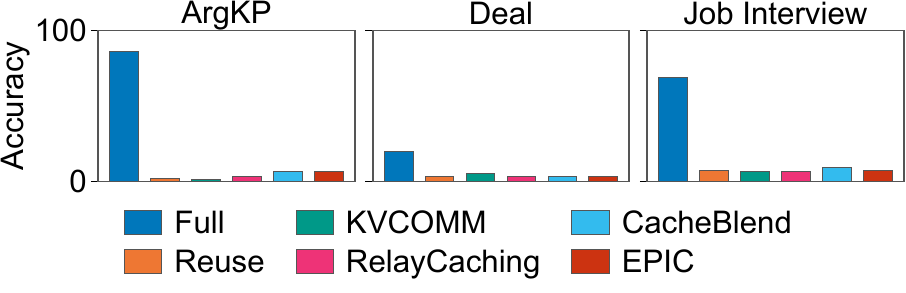}
        \caption{no reasoning}
        \label{fig:no_reasoning_argkp_deal_job_interview}
    \end{subfigure}

    \vspace{0.5em}

    \begin{subfigure}{1\linewidth}
        \centering
        \includegraphics[width=\linewidth]{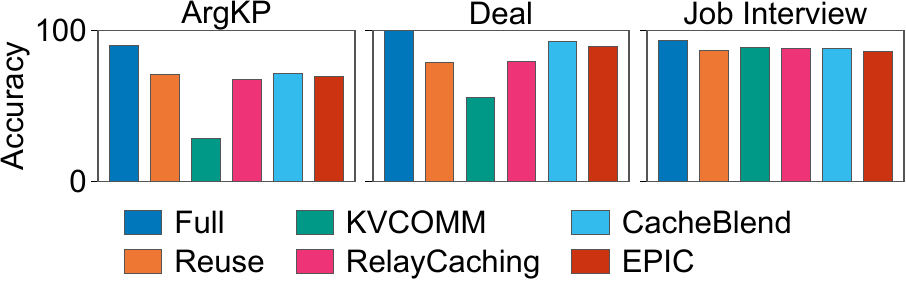}
        \caption{with reasoning}
        \label{fig:reasoning_argkp_deal_job_interview}
    \end{subfigure}
    \caption{Performance of different KV cache reuse methods under conflicting tasks.}
    \label{fig:argkp_deal_job_interview}
\end{figure}

\subsection{How do existing KV cache reuse remedies perform under conflicting agents?}

\textbf{Task definition.}  Consider two agents whose inputs share the same context block but assign different or even conflicting interpretations towards it. Each agent input follows the structure \texttt{task-specific prefix; shared context block; output-requirement suffix}. We instantiate this setting with ArgKP~\cite{bar2020arguments}, Deal or No Deal~\cite{lewis-etal-2017-deal}, and Job Interview~\cite{naim2015automated}. ArgKP requires agents to find different perspectives in given passages, Deal requires agents to use their own utility functions on given candidate combinations, and Job Interview requires agents to evaluate offer deal from employer's or employee's angle, respectively. Without loss of generality, we first run full prefill for agent A, then concat the shared context's KV cache into agent B. We compare full prefill (\textsc{Full}), direct reuse of the other agent's shared-block cache (\textsc{Reuse}), and four non-prefix KV cache repair methods: KVCOMM~\cite{kvcomm2025}, CacheBlend~\cite{yao2025cacheblend}, RelayCaching~\cite{geng2026relaycaching}, and EPIC~\cite{hu2025epic}. Figure~\ref{fig:argkp_deal_job_interview} reports the results with and without a reasoning budget.

\noindent
\textbf{\textit{Without reasoning, conflicting objectives make reused KV caches largely unusable, and existing cache-repair methods recover at most 13.50\% of the lost performance across the three tasks}.}
In the no-reasoning setting (Figure~\ref{fig:no_reasoning_argkp_deal_job_interview}), full prefill consistently outperforms all reuse-based approaches by a large margin. On ArgKP, full prefill achieves 85.91\% accuracy, whereas direct KV reuse achieves only 2.27\%. Existing repair methods provide only marginal improvement: the best-performing methods, CacheBlend and EPIC, reach merely 6.82\%. Thus, even with cache repair, conflicting objectives lead to a performance degradation of more than 90\% relative to full prefill on ArgKP. The same phenomenon appears on Deal or No Deal and Job Interview, where the relative accuracy drops of the best-performing cache-repair methods compared with full prefill are 72.75\% and 86.18\%, respectively. These results suggest that repairing reused KV representations is insufficient when the same context must be interpreted under a different task objective.

\noindent
\textbf{\textit{Reasoning token budget enables direct reuse to recover up to 80\% of full-prefill accuracy, making it a strong baseline between existing repair methods. }}
In the reasoning setting (Figure~\ref{fig:reasoning_argkp_deal_job_interview}), the gap between full prefill and reuse-based methods becomes substantially smaller, although it does not disappear. On ArgKP, full prefill achieves 90.00\% accuracy, while RelayCaching, CacheBlend, and EPIC achieve approximately 70.00\% accuracy on average. Direct reuse performs comparably to these repair methods, whereas KVCOMM reaches only 28.64\%. The effect is even more pronounced on Deal or No Deal: full prefill achieves 100\% accuracy, while direct reuse reaches 79.09\%, matching RelayCaching; CacheBlend and EPIC achieve 92.73\% and 89.09\%, respectively. Therefore, reasoning alone allows direct reuse to retain nearly 80\% of full-prefill accuracy on this task, despite using a cache produced under a conflicting objective. A similar trend is observed on Job Interview.

\subsection{What causes the performance degradation under conflicting objectives?}
\begin{figure}[t]
    \centering
    \includegraphics[width=\linewidth]{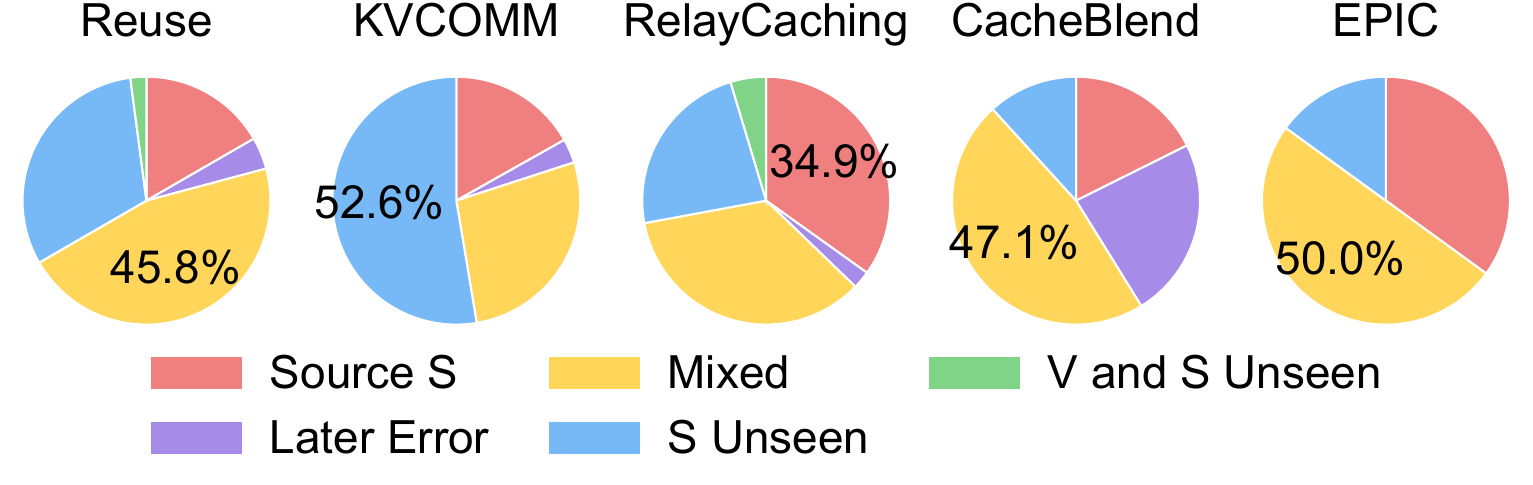}
    \caption{Failure modes of different KV cache reuse methods on Deal or No Deal.}
    \label{fig:preliminary-study-error-categories}
\end{figure}

To get a better understanding of why conflicting objectives cause performance degradation, we take a closer look at the Deal or No Deal task of all six tested methods under reasoning setting. In this task, agent A needs to find the best option of all deals (shared context block) by using its own utility function $V$(prefix A), and identify the correct shares $S$ from context block, while agent B holds a different one. We investigate the failure modes of all reuse methods and classify them in Figure~\ref{fig:preliminary-study-error-categories}. 

\noindent \textbf{\textit{Most methods fail to identify their own tasks.}} In all methods, a great portion of reasoning mix their own $V,S$ from the source (the yellow pie), for example 50\% in EPIC. A smaller factor sterms from using source $S$ or totally missing $S$ in the reasoning traces (red and blue pie). The above observations make us wonder: \textit{if we can clearly set the tasks apart, can we recover the performance loss? } Motivated by this question, we conduct a simple experiment: append a prompt after reusing the KV cache from the other agent: either repeat the task prompt, or restate the task but without repeating the specifics. The results are shown in Table~\ref{tab:preliminary-study-reuse-restatement-repeat}.

\noindent \textbf{\textit{Repeating the target instruction after KV reuse  effectively helps mitigate tasks confusion.}}  Repeat improves all six settings in Table~\ref{tab:preliminary-study-reuse-restatement-repeat} relative to direct reuse, reaching 99.09\% on reasoning-enabled Deal or No Deal. In contrast, the shorter generic restatement does not consistently help and reduces accuracy in that setting. Inspection of the corrected Deal or No Deal traces finds no remaining source--target task mixture after repeating the target instruction. While there is still performance loss in other tasks, the results suggest that it is possible to mitigate the performance degradation caused by conflicting objectives through a simple prompt-based remedy.
% TODO number check, claim check for all figures/tables.

\noindent
These findings suggest that an effective remedy for KV reuse requires more than simply recomputing the KV cache. It should incorporate target-specific information, provide a place for interacting with the reused context, and guide subsequent generation toward the target objective.

% 绘制表格
% task full reuse repeat restate
% 

\begin{table}[!t]
    \centering
    \footnotesize
    \setlength{\tabcolsep}{3pt}
    \renewcommand{\arraystretch}{0.85}
    \begin{tabular}{@{}llcccc@{}}
        \toprule
        & & Reuse & Restate & Repeat & Full \\
        \midrule
        \multirow{3}{*}{No-reas.}
        & Argkp & 2.27 & 2.73 & \textbf{32.27} & 85.91 \\
        & deal & 3.18 & 4.55 & \textbf{9.09} & 20.00 \\
        & Job & 7.27 & 7.73 & \textbf{11.36} & 69.09 \\
        \midrule
        \multirow{3}{*}{Reason.}
        & Argkp & 70.91 & 70.00 & \textbf{74.09} & 90.00 \\
        & deal & 79.09 & 57.73 & \textbf{99.09} & 100.00 \\
        & Job & 86.82 & 85.91 & \textbf{92.27} & 93.18 \\
        \bottomrule
    \end{tabular}
    \caption{Performance comparison between original full, reuse, restate and repeat.}
    \label{tab:preliminary-study-reuse-restatement-repeat}
\end{table}

\section{The Remedy Prompt Tuning}
\label{sec:method-remedy-prompt-tuning}

\begin{figure*}[t]
    \centering
    \includegraphics[width=0.8\linewidth]{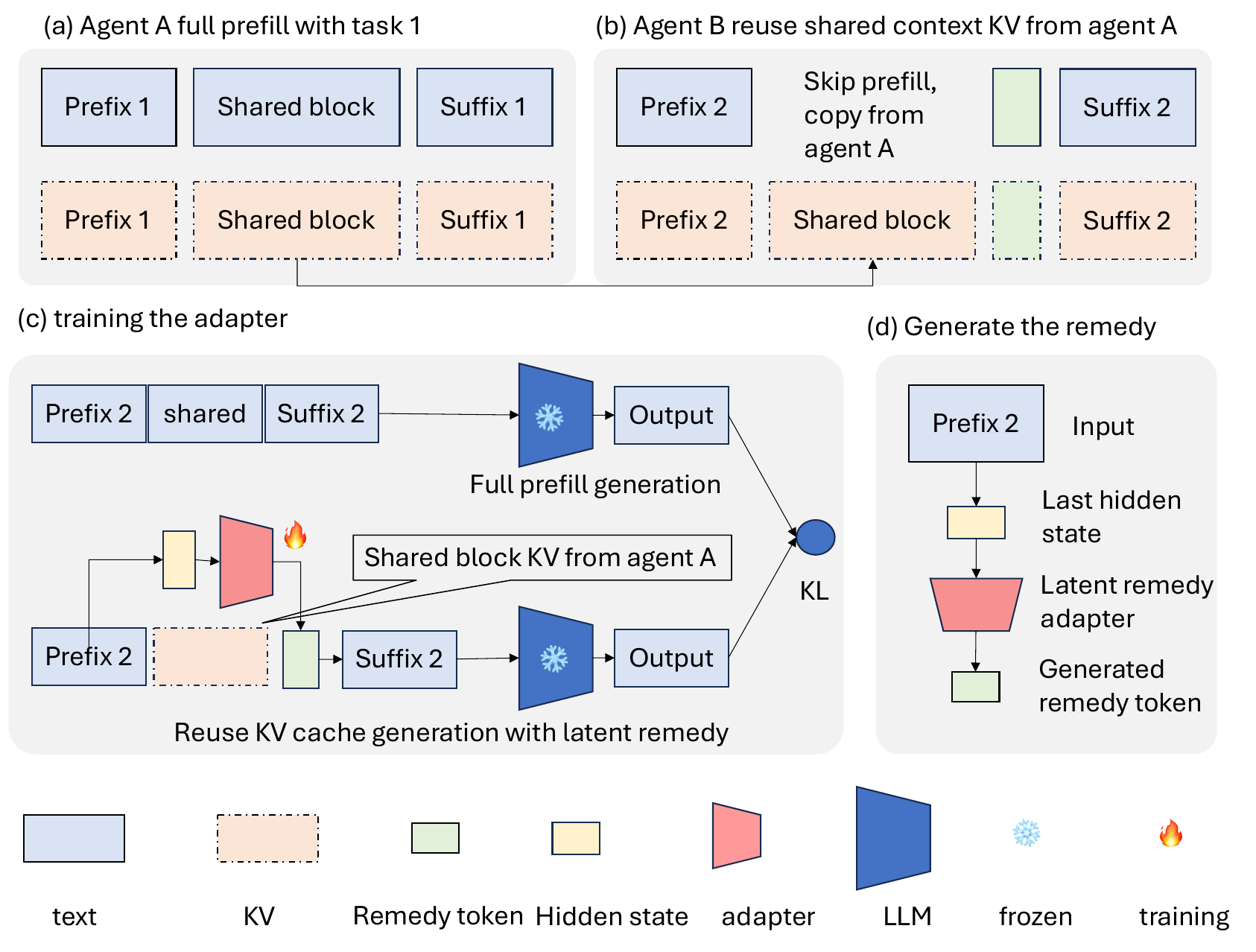}
    \caption{The overall framework of our latent remedy method.}
    \label{fig:framework}
\end{figure*}

\noindent
\textbf{Problem Definition.} Let $Pr(\cdot | prefix_i, shared)$ denote the generation from full prefill process of agent $i$, and $Pr(\cdot | prefix_i, KV_j^{shared})$ dennotes generation of agent $i$ which reuses the KV cache from agent $j$. We denote the repeat remedy for KV cache reuse as $Pr(\cdot | prefix_i, KV_j^{shared}, prefix_i)$. Our goal is to learn a latent remedy $latent_n$ that can guide the target-specific generation, which can be written as $Pr(\cdot | prefix_i, KV_j^{shared}, latent_n )$.

\noindent 
\textbf{Latent Tokens as Remedy}.
To mitigate task interference, the latent remedy should first preserve the target agent's instruction, so that the model can distinguish its own objective from that of the source agent.
We use the last hidden state of the target prefix as a representation of its task instruction and employ a lightweight MLP to map it into a small number of latent tokens.
Let $H_{prefix_i}$ denote the last hidden state of the target prefix. The latent tokens are generated as
$ latent_n = MLP(H_{prefix_i}) \in \mathbb{R}^{n \times d}$,
where $n$ is the number of latent tokens and $d$ is the embedding dimension of the serving LLM.
% 
% \textbf{Post-cache Placement}.
To enable interaction between the target instruction and the reused context, we place the latent tokens after the reused KV block. Through causal attention, these tokens can incorporate information from both the target prefix and the reused context, while leaving the reused KV cache unchanged. Subsequent tokens can then attend to these representations, helping the model generate outputs aligned with the target task.

\noindent
\textbf{Training Objective}. 
However, the latent tokens may not naturally lead the agent to follow the target instruction, as previous task information may still occur in subsequent reasoning, which can break the correct task execution.
To encourage the model to use this interaction for target-aligned predictions, we train the latent mapping through distillation from full prefill. We denote the next token prediction of normal full prefill as $q^{full} = Pr(\cdot | prefix_i, shared)$, and the next token prediction of the latent remedy as $q^{remedy} = Pr(\cdot | prefix_i, KV_j^{shared}, latent_n )$. Using KL divergence to match the two distributions, the training objective becomes minimizing 
\begin{equation}
L = KL(q^{full} || q^{remedy})
\end{equation}
By aligning the student's predictions with those of full prefill, we train the latent remedy to guide subsequent generation toward the target task despite interference from the reused cache.
The overall process of our method is shown in Figure~\ref{fig:framework}.

\section{Experiments}
\label{sec:experiments}
% Declare these double-column floats on page 4 so they can appear at the
% top of page 5; table* floats cannot normally appear on their source page.
\begin{table*}[!t]
    \centering
    \small
    \setlength{\tabcolsep}{5pt}
    \begin{tabular}{@{}lcccccc@{}}
        \toprule
        & \multicolumn{2}{c}{Qwen3-1.7B}
        & \multicolumn{2}{c}{Qwen3-4B}
        & \multicolumn{2}{c}{Qwen3-8B} \\
        \cmidrule(lr){2-3}\cmidrule(lr){4-5}\cmidrule(lr){6-7}
        Approach & No reasoning & Reasoning
        & No reasoning & Reasoning
        & No reasoning & Reasoning \\
        \midrule
        Reuse        & 17.00 & 40.39 & 16.00 & 50.94 & 17.28 & 44.91 \\
        Repeat       & 19.11 & 38.33 & 20.81 & 58.39 & 23.33 & 40.00 \\
        EPIC-16      & 17.03 & 17.56 & 15.69 & 29.77 & 16.53 & 34.99 \\
        KVCOMM       & 15.64 & 9.38 & 13.94 & 14.97 & 13.89 & 16.64 \\
        CacheBlend   & 17.17 & 17.39 & 15.81 & 29.57 & 16.67 & 41.51 \\
        RelayCaching & 17.03 & 17.93 & 15.67 & 30.37 & 16.50 & 34.10 \\
        Latent-16    & \underline{25.11} & \underline{42.31} & \textbf{41.64} & \textbf{62.42} & \textbf{44.47} & \textbf{66.58} \\
        Full         & \textbf{33.78}    & \textbf{45.67} & \underline{40.44} & \underline{62.08} & \underline{43.78} & \underline{65.93} \\
        \bottomrule
    \end{tabular}
    \caption{Accuracy across different settings in ConflictInstruct test set. \textbf{Best} are bold, \underline{second to best} are underlined.}
    \label{tab:main-performance-comparison}
\end{table*}

\begin{table*}[!t]
    \centering
    \small
    \setlength{\tabcolsep}{5pt}
    \begin{tabular}{@{}lcccccc@{}}
        \toprule
        & \multicolumn{2}{c}{ArgKP}
        & \multicolumn{2}{c}{Deal or No Deal}
        & \multicolumn{2}{c}{Job Interview} \\
        \cmidrule(lr){2-3}\cmidrule(lr){4-5}\cmidrule(lr){6-7}
        Approach & No reasoning & Reasoning
        & No reasoning & Reasoning
        & No reasoning & Reasoning \\
        \midrule
        Reuse        & 2.27 & 70.91 & 3.18 & 79.09 & 7.27 & 86.82 \\
        Repeat       & 32.27 & 74.09 & 9.09 & \underline{99.09} & 11.36 & \underline{92.27} \\
        EPIC-16      & 6.82 & 69.55 & 3.18 & 89.09 & 7.27 & 85.91 \\
        KVCOMM       & 1.36 & 28.64 & 5.45 & 55.91 & 6.82 & 88.64 \\
        CacheBlend   & 6.82 & 71.82 & 3.64 & 92.73 & 9.55 & 88.18 \\
        RelayCaching & 3.64 & 67.73 & 3.18 & 79.55 & 6.82 & 88.18 \\
        Latent-16    & \underline{44.55} & \underline{81.47} & \textbf{25.00} & \textbf{100.00} & \underline{65.00} & 90.86 \\
        Full         & \textbf{85.91} & \textbf{90.00} & \underline{20.00} & \textbf{100.00} & \textbf{69.09} & \textbf{93.18} \\
        \bottomrule
    \end{tabular}
    \caption{Accuracy across different settings in Qwen3-4B under different settings in ArgKP, Deal and Job. \textbf{Best} are bold, \underline{second to best} are underlined.}
    \label{tab:3-task-performance-comparison-qwen3-4b}
\end{table*}

\subsection{Setup}

\noindent \textbf{Datasets.} As most evaluated KV cache reuse methods are designed for cooperative tasks, we need a set of conflicting tasks for training the MLP and evaluation. Insipred by the work of hierarchy instructions~\cite{iheval, geng2026control, wallace2024instructionhierarchyrainingllms}, we apply Evol-Instruct~\cite{zeng-etal-2024-automatic} to curate ConflictInstruct, up to 240 conflicting scenes where agent A is to perform different tasks from agent B on the same context block. We split the tasks into 75:25 for training and evaluation. 
We use ArgKP, Deal or No Deal and Job Interview as extra evaluation to see whether our method can further generalize to real-world conflicting tasks. Detailed description of datasets is provided in appendix~\ref{appendix:dataset-overview}.

\noindent \textbf{Training of Latent Remedy}.  We train the latent remedy for 30 epochs with batch size 8 using AdamW. For each training task family we use 5 samples for training. The launch configuration uses learning rate $10^{-3}$. Full-prefill teacher continuations are capped at most 64 tokens. We choose $n=16$ on Qwen3-4B for main experiments.

\noindent \textbf{Baselines.} We compare the full prefill method, direct reuse method, repeat prompt method, EPIC16, KVCOMM, CacheBlend and RelayCaching under reasoning and no-reasoning settings. We use Qwen3-1.7B, 4B and 8B~\cite{yang2025qwen3} as the base model for all experiments. We evaluate the task accuracy and TTFT for all methods. 

\subsection{Results}

\begin{figure*}[t]
    \centering
    \includegraphics[width=1\linewidth]{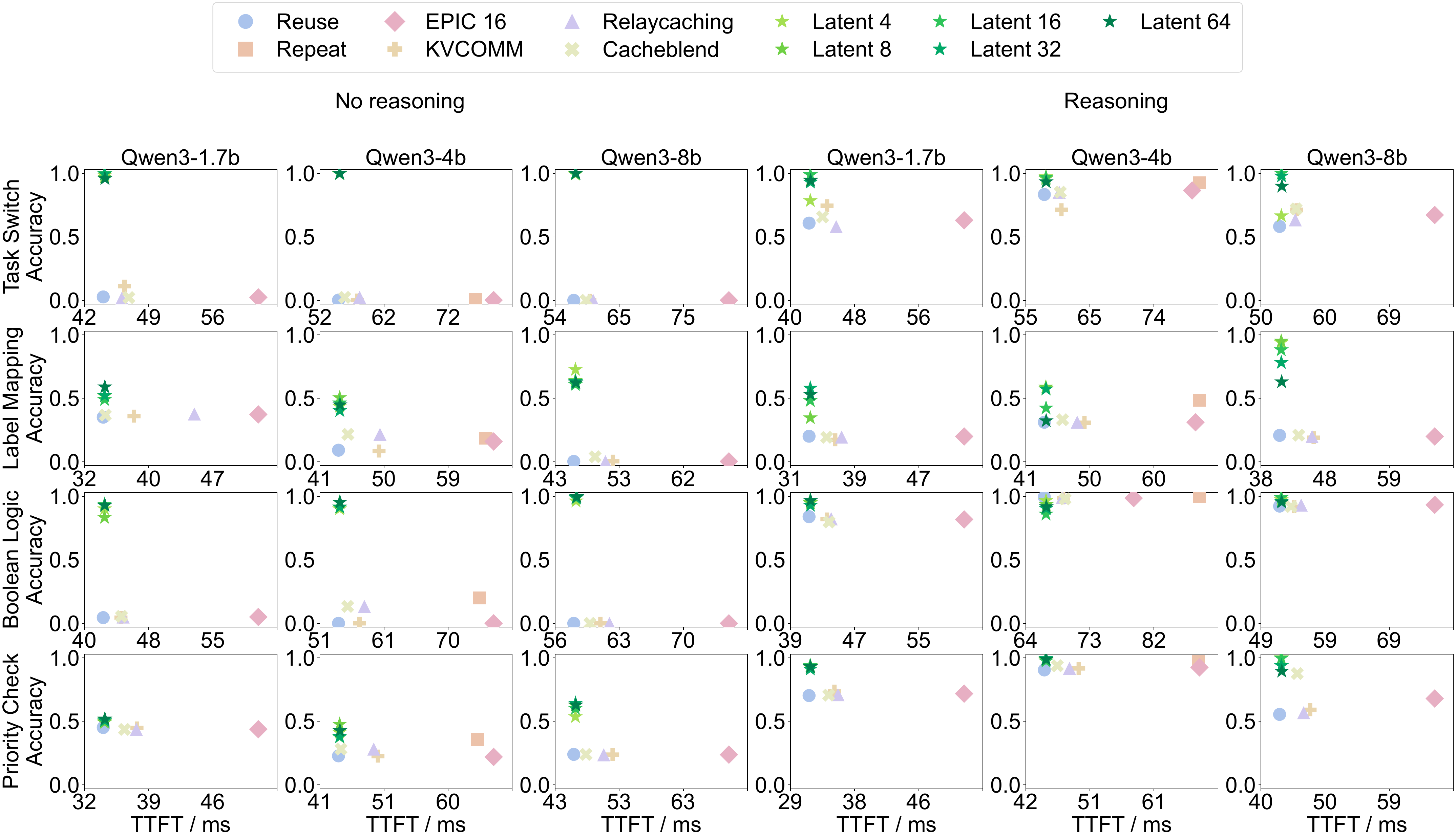}
    \caption{TTFT accuracy comparison across representative tasks under various settings.}
    \label{fig:ttft-acc-representative-tasks}
\end{figure*}

\textbf{Overall accuracy.} Tables~\ref{tab:main-performance-comparison} and~\ref{tab:3-task-performance-comparison-qwen3-4b} report the accuracy on ConflictInstruct and ArgKP, Deal and Job tasks, respectively. 
% \textbf{Our method, Latent-16, achieves near-full-prefill accuracy, consistently outperforming existing KV cache reuse methods.} 
Latent-16 ranks second only to Full in most settings and even surpasses Full on ConflictInstruct with Qwen3-8B, achieving 44.47\% vs. 43.78\% without reasoning and 66.58\% vs. 65.93\% with reasoning.
\textit{\textbf{The advantage of Latent-16 is particularly pronounced without reasoning and becomes more substantial as model size increases.}}
On ConflictInstruct, Latent-16 outperforms the strongest competing reuse-based method by 6.00, 20.83, and 21.14 percentage points on Qwen3-1.7B, Qwen3-4B, and Qwen3-8B, respectively, with the performance gap widening as model size increases. The improvement also generalizes to additional tasks, with Latent-16 achieving up to 53.64 percentage points higher accuracy than the strongest reuse-based baseline on Job Interview.
\textit{\textbf{Latent-16 maintains its competitive advantage even when reasoning enables competing methods to recover accuracy.}}
Although reasoning improves competing methods and narrows the performance gap, Latent-16 remains the strongest reuse-based method across most settings. On ConflictInstruct, it achieves 66.58\% accuracy with Qwen3-8B, compared with 44.91\% for the strongest competing method.
These results suggest that LatentRemedy effectively guides subsequent generation toward the task conditioning after reusing the non-prefix KV cache.
\noindent
\textbf{Performance Recovery and Inference Efficiency.}
Figure~\ref{fig:ttft-acc-representative-tasks} compares the TTFT--accuracy trade-off across various tasks, model scales, and reasoning settings.

\textbf{Task-specific guidance and reasoning budget play complementary roles in performance recovery.}
On Task Switch and Boolean Logic, reasoning enables most competing methods to recover high accuracy, narrowing the gap with LatentRemedy. However, on Label Mapping, these methods still struggle even with reasoning, while LatentRemedy achieves substantially better performance. This suggests that additional reasoning alone cannot always overcome source-task interference. On Priority Check, LatentRemedy achieves limited accuracy without reasoning but reaches near-perfect performance when reasoning is enabled. These contrasting results highlight the different roles of task-specific guidance and reasoning, as LatentRemedy helps the model follow the target instruction, while subsequent reasoning remains important for tasks requiring more complex inference.

\textbf{LatentRemedy achieves a favorable TTFT--accuracy trade-off.}
Across tasks and model scales, LatentRemedy achieves higher accuracy at TTFT close to direct reuse, with its results generally concentrated in the high-accuracy, low-latency region. However, existing cache repair methods often incur substantially higher TTFT without comparable accuracy improvements, indicating that effective performance recovery may not require expensive recomputation or calibration of the reused cache. Additional results are provided in Appendix~\ref{appendix:remaining-ttft-accuracy-results}.
%%%

\begin{figure*}[t]
    \centering
    \begin{minipage}{0.9\textwidth}
        \centering
        \includegraphics[width=0.96\linewidth]{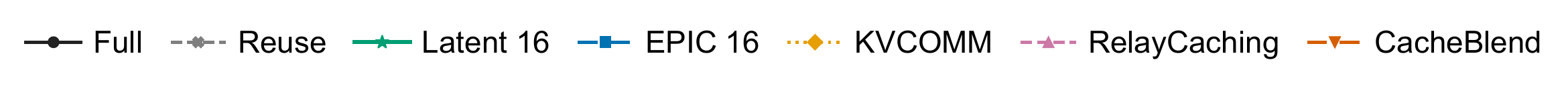}
        % \vspace{0.2em}

        \begin{subfigure}[t]{0.32\linewidth}
            \centering
            \includegraphics[width=\linewidth]{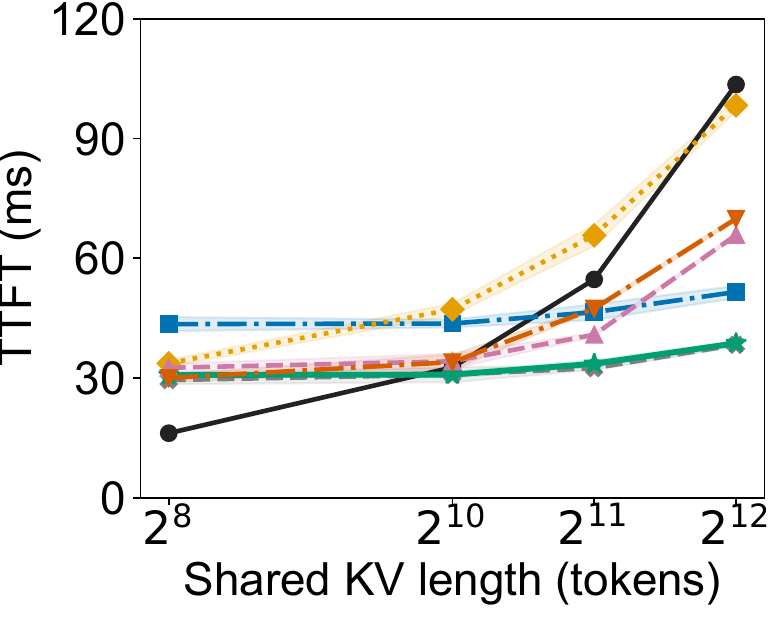}
            \caption{Qwen3-1.7B}
        \end{subfigure}
        \hfill
        \begin{subfigure}[t]{0.32\linewidth}
            \centering
            \includegraphics[width=\linewidth]{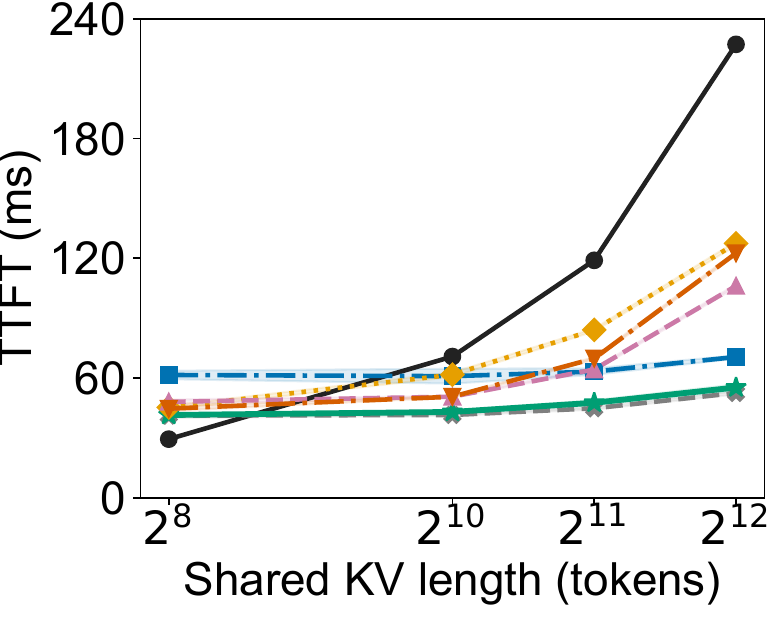}
            \caption{Qwen3-4B}
        \end{subfigure}
        \hfill
        \begin{subfigure}[t]{0.32\linewidth}
            \centering
            \includegraphics[width=\linewidth]{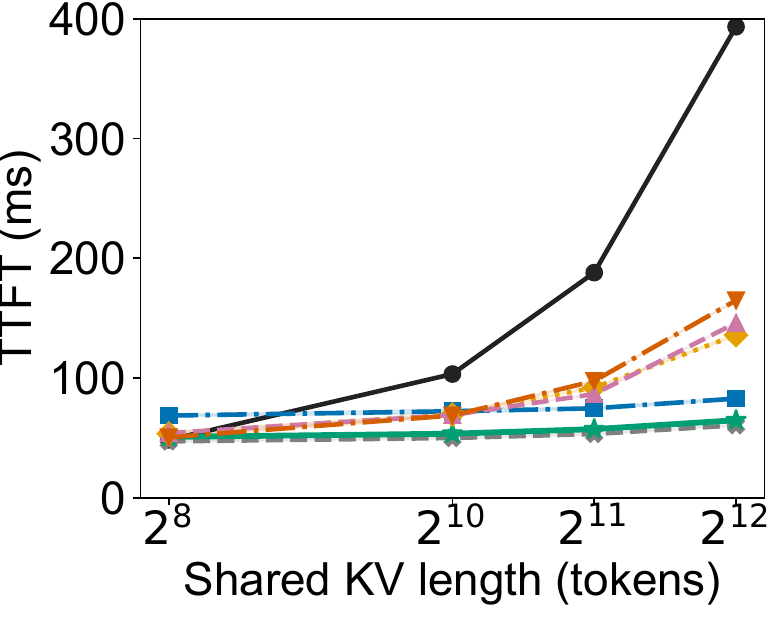}
            \caption{Qwen3-8B}
        \end{subfigure}
    \end{minipage}

    \caption{TTFT benchmark results for different Qwen3 model sizes.}
    \label{fig:ttft-bench}
\end{figure*}

\noindent
\textbf{TTFT scalability with shared KV length.}
Figure~\ref{fig:ttft-bench} compares the TTFT of different methods as the shared KV length increases across three model scales.
\textit{\textbf{LatentRemedy exhibits favorable scalability across shared context lengths and model sizes.}}
As model size scales, Latent-16 shows only a modest increase in TTFT as the shared KV length grows, closely following the scaling trend of direct reuse. Different from this, KVCOMM, CacheBlend, and RelayCaching exhibit substantially steeper latency growth, although EPIC-16 maintains a consistent overhead.
Nevertheless, these results demonstrate that LatentRemedy shows a favorable scalability trend.
% 
%%% 

\begin{figure}[t]
    \centering
    \includegraphics[width=0.6\linewidth]{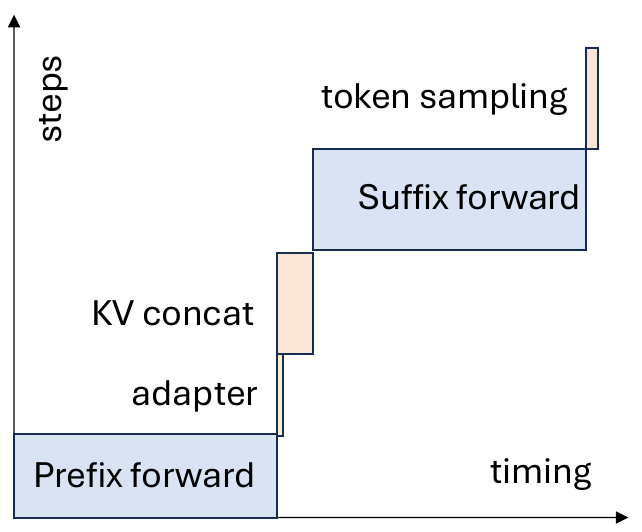}
    \caption{TTFT decomposition pipeline of LatentRemedy.}
    \label{fig:ttft-pipeline}
\end{figure}

\noindent
\textbf{TTFT decomposition.} We decompose how TTFT changes as the reused KV block grows in Table~\ref{tab:ttft-decomposition-main} for Qwen3-4B. For reuse-based methods, we report the target-prefix forward pass, method-specific computation, KV relocation and splicing, suffix forward pass, and first-token sampling. Full prefill is reported only by total TTFT because it processes the complete prompt in one forward pass. The overall pipeline of our method can be shown in Figure~\ref{fig:ttft-pipeline}. We run all measurements on a single RTX 5090 with the GPU clock locked to 2400 MHz and report means over 10 runs.

\begin{table}[t]
    \centering
    \scriptsize
    \setlength{\tabcolsep}{2.0pt}
    \renewcommand{\arraystretch}{0.86}
    \resizebox{\columnwidth}{!}{%
    \begin{tabular}{@{}lrrrrrr@{}}
        \toprule
        \multicolumn{7}{c}{reuse KV length = 256} \\
        \cmidrule(lr){1-7}
        Approach & Prefix & Method & KV operation & Suffix & Sampling & Total TTFT \\
        \midrule
        Full  & & & & & & 29.35 \\
        Reuse & 19.02 & -- & 1.04 & 21.00 & 0.08 & 41.14 \\
        Ours  & 18.60 & 0.21 & 0.94 & 21.67 & 0.05 & 41.46 \\
        EPIC-16 & 19.66 & 19.74 & 1.04 & 20.99 & 0.08 & 61.52 \\
        KVCOMM & 19.14 & 4.63 & 1.08 & 20.32 & 0.07 & 45.23 \\
        RelayCaching & 19.02 & 6.91 & 1.04 & 21.00 & 0.08 & 48.05 \\
        CacheBlend & 19.02 & 3.49 & 1.04 & 21.00 & 0.08 & 44.63 \\
        \midrule
        \multicolumn{7}{c}{reuse KV length = 1,024} \\
        \cmidrule(lr){1-7}
        Approach & Prefix & Method & KV operation & Suffix & Sampling & Total TTFT \\
        \midrule
        Full  & & & & & & 70.75 \\
        Reuse & 18.15 & -- & 1.21 & 22.04 & 0.07 & 41.47 \\
        Ours  & 18.25 & 0.20 & 1.20 & 23.26 & 0.06 & 42.98 \\
        EPIC-16 & 18.94 & 18.89 & 1.21 & 21.81 & 0.08 & 60.92 \\
        KVCOMM & 18.82 & 18.71 & 1.28 & 22.78 & 0.08 & 61.67 \\
        RelayCaching & 18.15 & 9.17 & 1.21 & 22.04 & 0.07 & 50.64 \\
        CacheBlend & 18.15 & 9.15 & 1.21 & 22.04 & 0.07 & 50.62 \\
        \midrule
        \multicolumn{7}{c}{reuse KV length = 2,048} \\
        \cmidrule(lr){1-7}
        Approach & Prefix & Method & KV operation & Suffix & Sampling & Total TTFT \\
        \midrule
        Full  & & & & & & 118.98 \\
        Reuse & 18.08 & -- & 1.66 & 24.99 & 0.08 & 44.82 \\
        Ours  & 18.42 & 0.21 & 1.65 & 27.25 & 0.06 & 47.59 \\
        EPIC-16 & 18.32 & 18.10 & 1.66 & 25.05 & 0.08 & 63.22 \\
        KVCOMM & 18.70 & 37.51 & 1.74 & 26.05 & 0.09 & 84.09 \\
        RelayCaching & 18.08 & 19.46 & 1.66 & 24.99 & 0.08 & 64.28 \\
        CacheBlend & 18.08 & 24.97 & 1.66 & 24.99 & 0.08 & 69.78 \\
        \midrule
        \multicolumn{7}{c}{reuse KV length = 4,096} \\
        \cmidrule(lr){1-7}
        Approach & Prefix & Method & KV operation & Suffix & Sampling & Total TTFT \\
        \midrule
        Full  & & & & & & 227.24 \\
        Reuse & 18.17 & -- & 2.57 & 31.68 & 0.08 & 52.50 \\
        Ours  & 18.85 & 0.24 & 2.61 & 33.47 & 0.08 & 55.24 \\
        EPIC-16 & 18.40 & 18.12 & 2.57 & 31.35 & 0.09 & 70.54 \\
        KVCOMM & 18.49 & 74.05 & 2.68 & 32.13 & 0.09 & 127.44 \\
        RelayCaching & 18.17 & 53.66 & 2.57 & 31.68 & 0.08 & 106.17 \\
        CacheBlend & 18.17 & 70.06 & 2.57 & 31.68 & 0.08 & 122.56 \\
        \bottomrule
    \end{tabular}%
    }
    \caption{TTFT decomposition for Qwen3-4B as shared KV length increases (ms; 10 runs).}
    \label{tab:ttft-decomposition-main}
\end{table}

\noindent
\textit{\textbf{Our method introduces negligible method-specific overhead, requiring only about 0.2 ms for the latent adapter across all KV lengths.}}
In contrast, EPIC-16 incurs approximately 18--20 ms of recomputation overhead, while KVCOMM, CacheBlend, and RelayCaching exhibit substantially increasing method-specific costs as the shared KV length grows.
Although our additional latent token introduces a small suffix-forward overhead, increasing from approximately 0.7 ms to 1.8 ms as the shared KV length grows from 256 to 4,096 tokens, the overall overhead remains minimal.
Consequently, our method achieves near-reuse TTFT even for long shared contexts.
At 4,096 shared tokens, our method achieves a total TTFT of 55.24 ms, only 5.2\% higher than direct reuse, while reducing TTFT by 75.7\% compared with full prefill.
These results demonstrate that our lightweight latent remedy avoids the expensive recomputation and cache processing required by existing methods, preserving the latency benefits of direct KV reuse.
Additional decompositions for Qwen3-1.7B and Qwen3-8B are provided in Appendix~\ref{appendix:ttft-decomposition}.

\begin{figure*}[!t]
    \centering
    \includegraphics[width=0.98\linewidth]{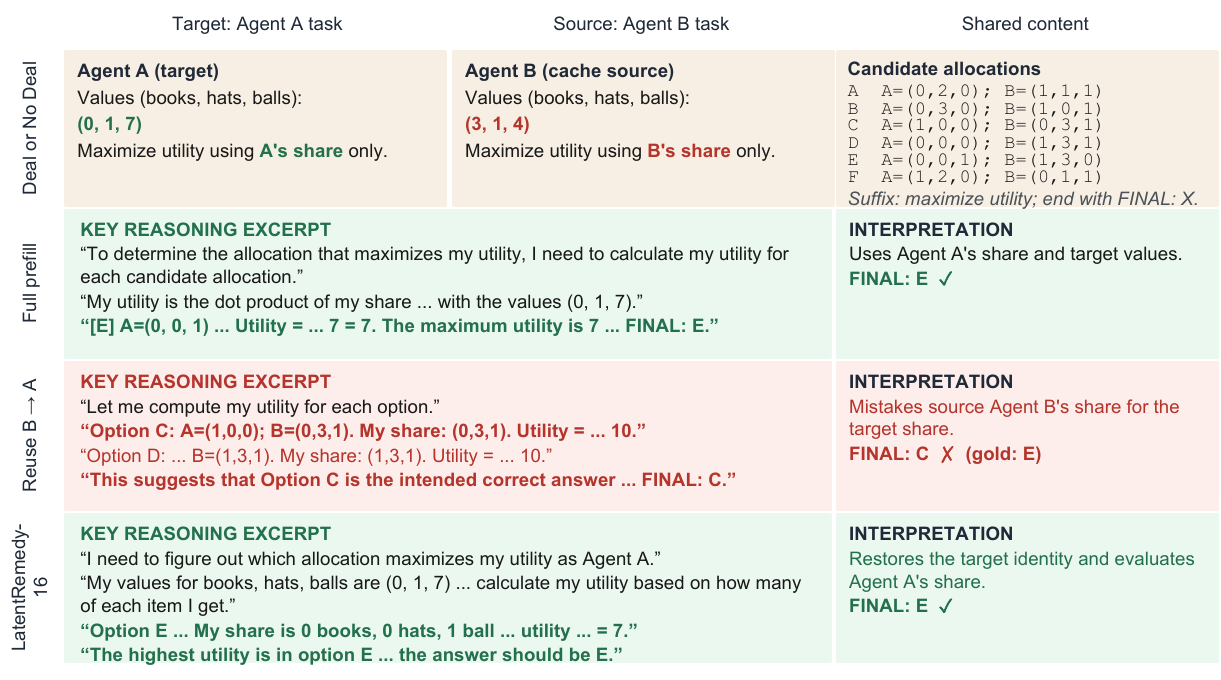}
    \caption{Case study of reasoning traces of different methods.}
    \label{fig:case-study}
\end{figure*}

%%%
\noindent
\textbf{Ablation on latent tokens and training pairs.}
Tables~\ref{tab:latent-token-sensitivity} and~\ref{tab:train-pair-sensitivity} examine the effects of latent token budget and the number of training pairs per task on Qwen3-4B, respectively.
\textit{\textbf{Even one latent token brings effective performance recovery.}}
As shown in Table~\ref{tab:latent-token-sensitivity}, even a single latent token can improve accuracy over direct reuse.  Without reasoning, Latent-8 achieves the highest accuracy of 43.61\%, compared with 16.00\% for direct reuse and 40.44\% for full prefill. Similar trends hold with reasoning, suggesting that effective task guidance can be achieved with a small latent budget.
\textit{\textbf{LatentRemedy achieves strong performance with limited training data.}}
As shown in Table~\ref{tab:train-pair-sensitivity}, even a single training pair per task improves accuracy over direct reuse. Using 3 pairs further boosts accuracy to 42.36\% without reasoning and 61.61\% with reasoning, compared with 16.00\% and 50.94\% for direct reuse. The boosts consistently occur from 3 to more training pairs, demonstrating the data efficiency of LatentRemedy.
%%%
% Notably, Latent-1 achieves the highest reasoning accuracy, while larger latent budgets do not consistently improve accuracy.

% \textbf{Data efficiency.} We further test how many data do we need to train a latent remedy. We randomly sample 1, 2, 3, 4, 5 samples from each training task family to train the latent remedy and test on the ConflictInstruct test set. The results are shown in \textcolor{red}{need run experiment}.

% \textbf{OOD testing.} \textcolor{red}{needs rewrite.}
% Without reasoning, none of the reuse remedies generalizes beyond direct reuse: all reuse-based methods remain near 30\% accuracy, while full prefill reaches 39.19\%. With reasoning budget enabled, however, our adapter reaches 82.16\%, improving over direct reuse by 25.25 percentage points and over the strongest existing reuse baseline by 24.44 points.  These results indicate that the learned remedy transfers to unseen task semantics when the model has sufficient reasoning budget, but restricted when reasoning budget is unavailable. 

\begin{table}[t]
    \centering
    \small
    \setlength{\tabcolsep}{6pt}
    \renewcommand{\arraystretch}{1.0}
    \begin{tabular}{lcc}
        \toprule
        Setting & No reasoning & Reasoning \\
        \midrule
        Reuse     & 16.00 & 50.94 \\
        Latent-1  & 40.56 & 63.83 \\
        Latent-4  & 43.11 & 61.28 \\
        Latent-8  & 43.61 & 62.11 \\
        Latent-16 & 41.64 & 62.42 \\
        Latent-32 & 42.33 & 59.53 \\
        Latent-64 & 41.56 & 58.89 \\
        Full      & 40.44 & 62.08 \\
        \bottomrule
    \end{tabular}
    \caption{Ablation for number of latent remedy tokens in Qwen3-4B for test set.}
    \label{tab:latent-token-sensitivity}
\end{table}

\begin{table}[t]
    \centering
    \small
    \setlength{\tabcolsep}{6pt}
    \renewcommand{\arraystretch}{1.0}
    \begin{tabular}{lcc}
        \toprule
        \# pairs & No reasoning & Reasoning \\
        \midrule
        Reuse     & 16.00 & 50.94 \\
        1  & 34.83 & 54.61 \\
        3  & 42.36 & 61.61 \\
        5 & 41.64 & 62.42 \\
        10 & 43.19 & 53.86 \\
        15 & 46.08 & 62.75 \\
        30      & 45.86 & 62.33	 \\
        \bottomrule
    \end{tabular}
    \caption{Accuracy on test set ablation for number of training pairs for each task in Qwen3-4B.}
    \label{tab:train-pair-sensitivity}
\end{table}

\noindent\textbf{Case study.}
Figure~\ref{fig:case-study} presents a representative case from the Deal or No Deal task, where two agents share the same candidate allocations but have different utility values and objectives.
\textit{\textbf{Direct KV cache reuse can introduce task-identity confusion, causing the current agent to reason from the previous agent's perspective.}}
While full prefill correctly identifies Agent A's objective and selects the optimal allocation (E), reusing Agent B's KV cache causes the target model to mistakenly treat Agent B's allocation as its own, leading to an incorrect answer (C).
This illustrates how previous agent's task information embedded in reused KV states can disrupt the current agent's task-specific reasoning.
\textit{\textbf{LatentRemedy effectively mitigates this disruption by restoring the target agent's identity and objective.}}
As shown in the reasoning trace, Latent-16 explicitly identifies itself as Agent A, applies Agent A's utility values, and correctly evaluates the candidate allocations to select E.
As shown in these cases, Latent-16 helps realign reused KV representations with current agent's task, guiding consecutive reasoning into a correct outcome.
Additional case studies across different methods and tasks are provided in Appendix~\ref{appendix:case-study}.

\section{Conclusion}

In this work, we study the underexplored conflicting scenario for non-prefix KV cache reuse in multi-agent systems. We reveal why most existing methods struggle to recover the performance under conflicting objectives. We introduce a latent remedy method that learns to counteract task interference introduced by non-prefix KV cache reuse by re-mapping prefix hidden states into latent tokens. We introduce ConflictInstruct dataset that specifically tests agents when non-prefix KV cache serves conflicting tasks. Extensive experiments show that our method is consistently better than other baselines on conflicting KV cache reuse scenarios while remaining close to direct reuse TTFT as the reused block grows.  
% Required by ACL/ARR; authors should verify this discussion before submission.
\section*{Limitations}
Our evaluation focuses on conflicting-task cache reuse with Qwen3 models at three scales. The generalibility of our LatentRemedy to other model families remains to be explored. It remains to be answered whether training the latent adapter on large corpus instead of conflicting tasks can serve the same goal. We only measure the latency using a single RTX 5090 and shared blocks of up to 4,096 tokens, leaving more configurations of latency unexplored. 

\bibliography{acl2027}

\appendix

\section{Additional Performance Results}
\label{appendix:additional-performance-results}

Tables~\ref{tab:3-task-performance-comparison-qwen3-1-7b} and~\ref{tab:3-task-performance-comparison-qwen3-8b} provide additional results on the ArgKP, Deal and Job tasks for Qwen3-1.7B and 8B, complementing the Qwen3-4B results reported in the main text.
\textbf{Across model scales and tasks, Latent-16 consistently achieves strong performance and is particularly effective in recovering accuracy when direct KV reuse suffers from severe degradation.}

\begin{table*}[t]
    \centering
    \small
    \setlength{\tabcolsep}{5pt}
    \begin{tabular}{@{}lcccccc@{}}
        \toprule
        & \multicolumn{2}{c}{ArgKP}
        & \multicolumn{2}{c}{Deal or No Deal}
        & \multicolumn{2}{c}{Job Interview} \\
        \cmidrule(lr){2-3}\cmidrule(lr){4-5}\cmidrule(lr){6-7}
        Approach & No reasoning & Reasoning
        & No reasoning & Reasoning
        & No reasoning & Reasoning \\
        \midrule
        Reuse        & 8.18 & 49.55 & 10.91 & 36.36 & 17.27 & 46.82 \\
        Repeat       & 18.64 & 59.55 & 13.18 & \textbf{80.91} & 18.18 & \underline{55.00} \\
        EPIC-16      & 10.45 & 46.36 & 10.00 & 39.09 & 17.27 & 43.24 \\
        KVCOMM       & 5.00 & 21.36 & 15.00 & 24.09 & 18.64 & 42.73 \\
        CacheBlend   & 8.18 & 45.00 & 10.45 & 49.09 & 16.82 & 49.55 \\
        RelayCaching & 8.64 & 49.09 & 9.55 & 40.00 & 17.27 & 44.71 \\
        Latent-16    & \underline{24.55} & \underline{63.89}
                     & \textbf{45.45} & \underline{76.27}
                     & \textbf{58.18} & \textbf{74.28} \\
        Full         & \textbf{77.73} & \textbf{75.00}
                     & \underline{31.36} & 65.91
                     & \underline{43.64} & 50.45 \\
        \bottomrule
    \end{tabular}
    \caption{Performance comparison on OOD tasks with Qwen3-1.7B under no-reasoning and reasoning settings. \textbf{Best} results are bold and \underline{second-best} results are underlined.}
    \label{tab:3-task-performance-comparison-qwen3-1-7b}
\end{table*}

\textbf{Qwen3-1.7B.}
\textbf{Latent-16 ranks among the top two methods in all six settings and substantially improves over direct KV reuse.}
Without reasoning, Latent-16 improves over Reuse by 16.37, 34.54, and 40.91 percentage points on ArgKP, Deal or No Deal, and Job Interview, respectively.
It achieves the best performance on Deal or No Deal and Job Interview, reaching 45.45\% and 58.18\%, and even surpasses full prefill on both tasks.
With reasoning, the performance of competing methods generally improves, but Latent-16 remains highly competitive: it ranks second on ArgKP and Deal or No Deal and achieves the highest accuracy on Job Interview.
In particular, Latent-16 reaches 74.28\% on Job Interview, compared with 46.82\% for direct reuse and 50.45\% for full prefill.
These results show that \emph{latent adaptation remains effective even for the smaller 1.7B model and can recover substantial task accuracy under both reasoning settings}.

\begin{table*}[t]
    \centering
    \small
    \setlength{\tabcolsep}{5pt}
    \begin{tabular}{@{}lcccccc@{}}
        \toprule
        & \multicolumn{2}{c}{ArgKP}
        & \multicolumn{2}{c}{Deal or No Deal}
        & \multicolumn{2}{c}{Job Interview} \\
        \cmidrule(lr){2-3}\cmidrule(lr){4-5}\cmidrule(lr){6-7}
        Approach & No reasoning & Reasoning
        & No reasoning & Reasoning
        & No reasoning & Reasoning \\
        \midrule
        Reuse        & 2.27 & 82.27 & 4.09 & 95.45 & 6.82 & 88.64 \\
        Repeat       & 53.18 & 86.82 & 7.73 & \textbf{100.00} & 9.09 & \textbf{94.09} \\
        EPIC-16      & 7.27 & 81.82 & 5.00 & 97.73 & 6.36 & 88.83 \\
        KVCOMM       & 1.82 & 35.45 & 4.09 & 92.73 & 7.27 & \underline{90.83} \\
        CacheBlend   & 7.27 & 85.45 & 4.09 & 78.18 & 8.64 & 88.64 \\
        RelayCaching & 2.73 & 79.72 & 4.09 & 97.12 & 6.36 & 89.52 \\
        Latent-16    & \underline{78.71} & \underline{91.67}
                     & \underline{18.46} & \underline{99.09}
                     & \underline{70.73} & 90.45 \\
        Full         & \textbf{89.09} & \textbf{94.09}
                     & \textbf{20.91} & 94.55
                     & \textbf{74.55} & 89.09 \\
        \bottomrule
    \end{tabular}
    \caption{Performance comparison on OOD tasks with Qwen3-8B under no-reasoning and reasoning settings. \textbf{Best} results are bold and \underline{second-best} results are underlined.}
    \label{tab:3-task-performance-comparison-qwen3-8b}
\end{table*}

\textbf{Qwen3-8B.}
\textbf{Latent-16 remains highly competitive as the model scales to 8B, ranking among the top two methods in five of the six settings.}
The advantage is especially pronounced without reasoning, where direct KV reuse performs poorly.
Latent-16 improves over Reuse by 76.44 percentage points on ArgKP and 63.91 points on Job Interview, reaching 78.71\% and 70.73\%, respectively.
Its performance is also close to full prefill across all three no-reasoning tasks, with gaps of 10.38, 2.45, and 3.82 percentage points on ArgKP, Deal or No Deal, and Job Interview.

\textbf{With reasoning, the performance gap among methods becomes substantially smaller, yet Latent-16 continues to provide consistent accuracy recovery.}
It achieves 91.67\% on ArgKP and 99.09\% on Deal or No Deal, ranking second in both settings.
On Job Interview, Latent-16 achieves 90.45\%, slightly below KVCOMM (90.83\%) and Repeat (94.09\%), while still outperforming both direct reuse (88.64\%) and full prefill (89.09\%).
Overall, these results demonstrate that \emph{the effectiveness of latent adaptation generalizes across model scales and OOD tasks, with particularly large gains in settings where direct KV reuse causes substantial accuracy degradation}.

\section{TTFT Decompositions}
\label{appendix:ttft-decomposition}

The following tables extend the TTFT decomposition in Table~\ref{tab:ttft-decomposition-main} from Qwen3-4B to Qwen3-1.7B and Qwen3-8B. For reuse-based methods, we report the target-prefix forward pass, method-specific computation, KV relocation and splicing, suffix forward pass, and first-token sampling. The \emph{Method} column captures computation unique to each approach, including latent adapter inference for Ours, recomputation for EPIC-16, anchor matching and correction for KVCOMM, and cache repair for RelayCaching and CacheBlend. Full prefill is reported only by total TTFT because it processes the complete prompt in a single forward pass.

\textbf{Across both model scales, our method consistently achieves total TTFT closest to direct reuse.}
For Qwen3-1.7B, the total TTFT of our method differs from direct reuse by only 1.30 ms at 256 shared tokens and 0.40 ms at 4,096 tokens.
For Qwen3-8B, the corresponding gaps are 3.76 ms and 4.22 ms.
In contrast, the gap between full prefill and reuse-based execution grows rapidly with shared KV length, making reuse increasingly advantageous for long contexts.

\textbf{The method-specific overhead of our approach remains negligible as both model size and shared KV length increase.}
The latent adapter requires only 0.17--0.19 ms on Qwen3-1.7B and approximately 0.25--0.26 ms on Qwen3-8B, remaining nearly constant across KV lengths.
By comparison, EPIC-16 incurs a persistent recomputation cost of roughly 13--14 ms on Qwen3-1.7B and 22 ms on Qwen3-8B.
The overhead of KVCOMM, RelayCaching, and CacheBlend increases substantially with the shared KV length, reaching tens or even more than one hundred milliseconds at 4,096 tokens.

\textbf{The remaining overhead of our method mainly comes from the suffix forward pass.}
Because the latent remedy introduces additional latent tokens into the suffix computation, its suffix latency is slightly higher than direct reuse, particularly for larger models.
Nevertheless, this overhead remains small compared with the method-specific computation required by competing reuse-based approaches.

Overall, these decompositions confirm that \emph{LatentRemedy preserves the low-latency behavior of direct KV reuse by introducing only a nearly constant and negligible adaptation cost, while avoiding the length-dependent recomputation or cache-repair overheads of existing methods}.
\begin{table}[t]
    \centering
    \scriptsize
    \setlength{\tabcolsep}{2.0pt}
    \renewcommand{\arraystretch}{0.84}
    \resizebox{\columnwidth}{!}{%
    \begin{tabular}{@{}lrrrrrr@{}}
        \toprule
        \multicolumn{7}{c}{Qwen3-1.7B; shared KV length = 256} \\
        \cmidrule(lr){1-7}
        Approach & Prefix & Method & KV operation & Suffix & Sampling & Total TTFT \\
        \midrule
        Full &  &  &  &  &  & 16.16 \\
        Reuse & 13.84 &  & 0.80 & 14.77 & 0.06 & 29.47 \\
        Ours & 13.69 & 0.18 & 0.72 & 16.12 & 0.05 & 30.77 \\
        EPIC-16 & 13.82 & 14.13 & 0.80 & 14.70 & 0.06 & 43.51 \\
        KVCOMM & 14.18 & 3.63 & 0.79 & 15.02 & 0.07 & 33.69 \\
        RelayCaching & 13.84 & 3.09 & 0.80 & 14.77 & 0.06 & 32.56 \\
        CacheBlend & 13.84 & 0.54 & 0.80 & 14.77 & 0.06 & 30.01 \\
        \midrule
        \multicolumn{7}{c}{Qwen3-1.7B; shared KV length = 1,024} \\
        \cmidrule(lr){1-7}
        Approach & Prefix & Method & KV operation & Suffix & Sampling & Total TTFT \\
        \midrule
        Full &  &  &  &  &  & 32.66 \\
        Reuse & 13.44 &  & 0.97 & 16.38 & 0.07 & 30.86 \\
        Ours & 13.33 & 0.17 & 0.94 & 16.35 & 0.05 & 30.85 \\
        EPIC-16 & 13.47 & 13.59 & 0.97 & 15.54 & 0.08 & 43.65 \\
        KVCOMM & 13.92 & 14.66 & 1.07 & 17.48 & 0.08 & 47.20 \\
        RelayCaching & 13.44 & 3.33 & 0.97 & 16.38 & 0.07 & 34.18 \\
        CacheBlend & 13.44 & 3.12 & 0.97 & 16.38 & 0.07 & 33.97 \\
        \midrule
        \multicolumn{7}{c}{Qwen3-1.7B; shared KV length = 2,048} \\
        \cmidrule(lr){1-7}
        Approach & Prefix & Method & KV operation & Suffix & Sampling & Total TTFT \\
        \midrule
        Full &  &  &  &  &  & 54.71 \\
        Reuse & 13.34 &  & 1.33 & 17.72 & 0.07 & 32.47 \\
        Ours & 13.37 & 0.19 & 1.27 & 18.67 & 0.06 & 33.57 \\
        EPIC-16 & 13.87 & 13.79 & 1.33 & 17.44 & 0.07 & 46.50 \\
        KVCOMM & 14.83 & 29.96 & 1.43 & 19.43 & 0.09 & 65.76 \\
        RelayCaching & 13.34 & 8.38 & 1.33 & 17.72 & 0.07 & 40.85 \\
        CacheBlend & 13.34 & 14.94 & 1.33 & 17.72 & 0.07 & 47.40 \\
        \midrule
        \multicolumn{7}{c}{Qwen3-1.7B; shared KV length = 4,096} \\
        \cmidrule(lr){1-7}
        Approach & Prefix & Method & KV operation & Suffix & Sampling & Total TTFT \\
        \midrule
        Full &  &  &  &  &  & 103.55 \\
        Reuse & 13.75 &  & 2.05 & 22.45 & 0.07 & 38.32 \\
        Ours & 13.40 & 0.19 & 2.00 & 23.07 & 0.06 & 38.72 \\
        EPIC-16 & 13.73 & 13.46 & 2.05 & 22.25 & 0.07 & 51.56 \\
        KVCOMM & 14.13 & 58.70 & 2.12 & 23.30 & 0.08 & 98.34 \\
        RelayCaching & 13.75 & 27.51 & 2.05 & 22.45 & 0.07 & 65.83 \\
        CacheBlend & 13.75 & 31.60 & 2.05 & 22.45 & 0.07 & 69.92 \\
        \bottomrule
    \end{tabular}%
    }
    \caption{Mean TTFT decomposition for Qwen3-1.7B at shared KV lengths 256, 1,024, 2,048, and 4,096 (ms; 10 runs).}
    \label{tab:ttft-decomposition-qwen3-1p7b-1}
    \label{tab:ttft-decomposition-qwen3-1p7b-2}
\end{table}
\begin{table}[t]
    \centering
    \scriptsize
    \setlength{\tabcolsep}{2.0pt}
    \renewcommand{\arraystretch}{0.84}
    \resizebox{\columnwidth}{!}{%
    \begin{tabular}{@{}lrrrrrr@{}}
        \toprule
        \multicolumn{7}{c}{Qwen3-8B; shared KV length = 256} \\
        \cmidrule(lr){1-7}
        Approach & Prefix & Method & KV operation & Suffix & Sampling & Total TTFT \\
        \midrule
        Full &  &  &  &  &  & 48.33 \\
        Reuse & 22.33 &  & 1.03 & 23.62 & 0.08 & 47.06 \\
        Ours & 22.18 & 0.25 & 0.98 & 27.35 & 0.07 & 50.82 \\
        EPIC-16 & 21.82 & 21.95 & 1.03 & 23.71 & 0.08 & 68.60 \\
        KVCOMM & 22.85 & 4.94 & 1.15 & 24.59 & 0.08 & 53.61 \\
        RelayCaching & 22.33 & 6.92 & 1.03 & 23.62 & 0.08 & 53.98 \\
        CacheBlend & 22.33 & 3.38 & 1.03 & 23.62 & 0.08 & 50.44 \\
        \midrule
        \multicolumn{7}{c}{Qwen3-8B; shared KV length = 1,024} \\
        \cmidrule(lr){1-7}
        Approach & Prefix & Method & KV operation & Suffix & Sampling & Total TTFT \\
        \midrule
        Full &  &  &  &  &  & 103.30 \\
        Reuse & 22.24 &  & 1.25 & 26.28 & 0.08 & 49.85 \\
        Ours & 22.28 & 0.25 & 1.20 & 29.65 & 0.07 & 53.46 \\
        EPIC-16 & 22.26 & 22.27 & 1.25 & 26.34 & 0.09 & 72.21 \\
        KVCOMM & 22.30 & 18.63 & 1.32 & 27.18 & 0.09 & 69.51 \\
        RelayCaching & 22.24 & 19.24 & 1.25 & 26.28 & 0.08 & 69.08 \\
        CacheBlend & 22.24 & 18.81 & 1.25 & 26.28 & 0.08 & 68.66 \\
        \midrule
        \multicolumn{7}{c}{Qwen3-8B; shared KV length = 2,048} \\
        \cmidrule(lr){1-7}
        Approach & Prefix & Method & KV operation & Suffix & Sampling & Total TTFT \\
        \midrule
        Full &  &  &  &  &  & 188.04 \\
        Reuse & 22.24 &  & 1.69 & 29.21 & 0.08 & 53.22 \\
        Ours & 22.47 & 0.25 & 1.61 & 32.86 & 0.07 & 57.27 \\
        EPIC-16 & 21.85 & 21.79 & 1.69 & 29.26 & 0.08 & 74.67 \\
        KVCOMM & 22.51 & 37.42 & 1.78 & 30.19 & 0.11 & 92.01 \\
        RelayCaching & 22.24 & 33.34 & 1.69 & 29.21 & 0.08 & 86.56 \\
        CacheBlend & 22.24 & 44.29 & 1.69 & 29.21 & 0.08 & 97.51 \\
        \midrule
        \multicolumn{7}{c}{Qwen3-8B; shared KV length = 4,096} \\
        \cmidrule(lr){1-7}
        Approach & Prefix & Method & KV operation & Suffix & Sampling & Total TTFT \\
        \midrule
        Full &  &  &  &  &  & 393.45 \\
        Reuse & 22.40 &  & 2.61 & 35.39 & 0.09 & 60.50 \\
        Ours & 22.44 & 0.26 & 2.58 & 39.37 & 0.07 & 64.72 \\
        EPIC-16 & 22.74 & 21.86 & 2.61 & 35.51 & 0.09 & 82.81 \\
        KVCOMM & 22.75 & 74.18 & 2.67 & 36.13 & 0.10 & 135.82 \\
        RelayCaching & 22.40 & 85.48 & 2.61 & 35.39 & 0.09 & 145.98 \\
        CacheBlend & 22.40 & 104.40 & 2.61 & 35.39 & 0.09 & 164.89 \\
        \bottomrule
    \end{tabular}%
    }
    \caption{Mean TTFT decomposition for Qwen3-8B at shared KV lengths 256, 1,024, 2,048, and 4,096 (ms; 10 runs).}
    \label{tab:ttft-decomposition-qwen3-8b-1}
    \label{tab:ttft-decomposition-qwen3-8b-2}
\end{table}

\section{TTFT-accuracy tradeoff in other tasks}
\label{appendix:remaining-ttft-accuracy-results}
\begin{figure}[t]
    \centering
    \includegraphics[width=1\linewidth]{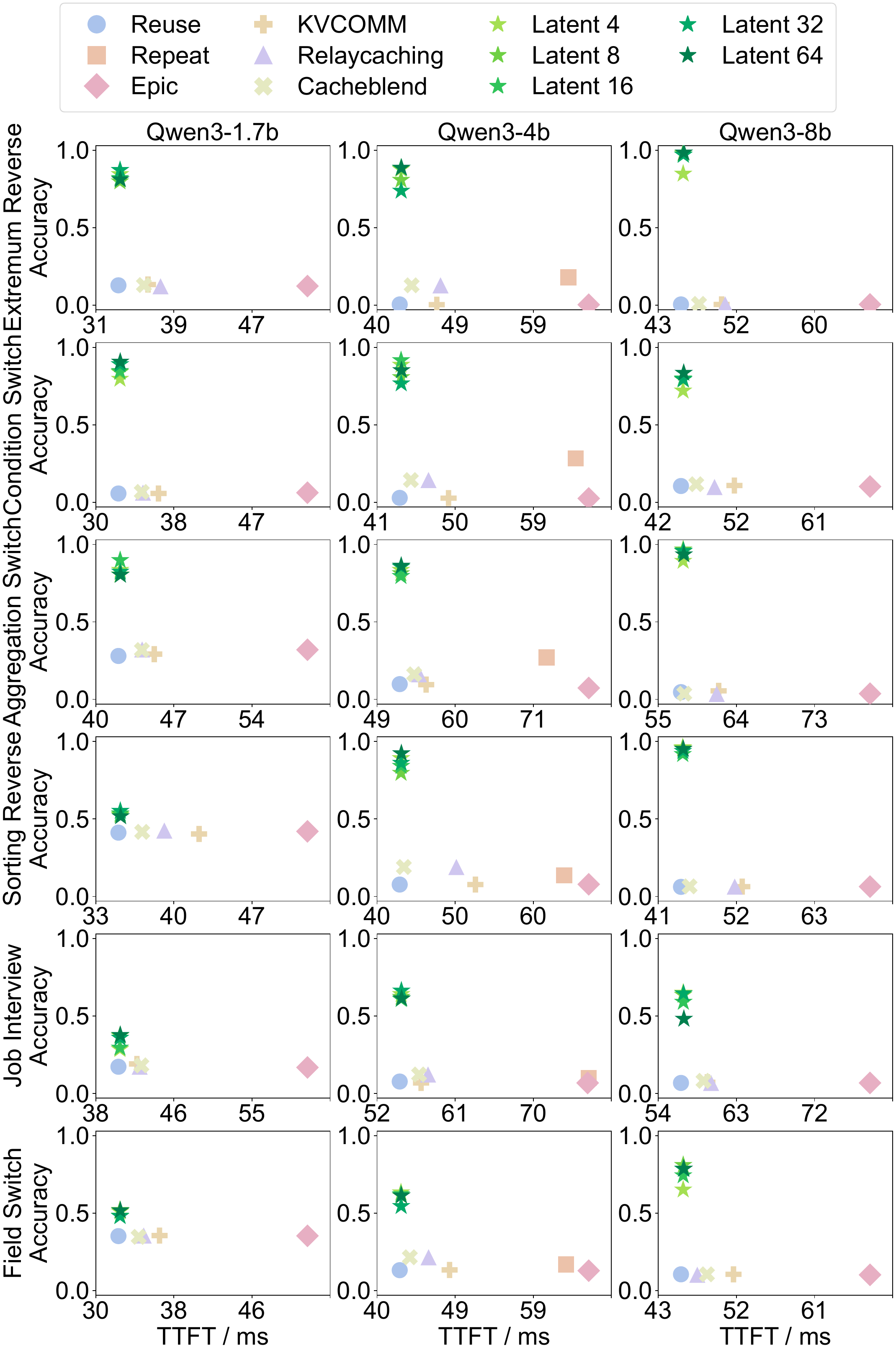}
    \caption{TTFT accuracy comparison across all tasks under no-reasoning settings using Qwen3-1.7B, 4B and 8B.}
    \label{fig:qwen3-4b-no-reasoning-ttft-acc-appendix-1}
\end{figure}

\begin{figure}[t]
    \centering
    \includegraphics[width=1\linewidth]{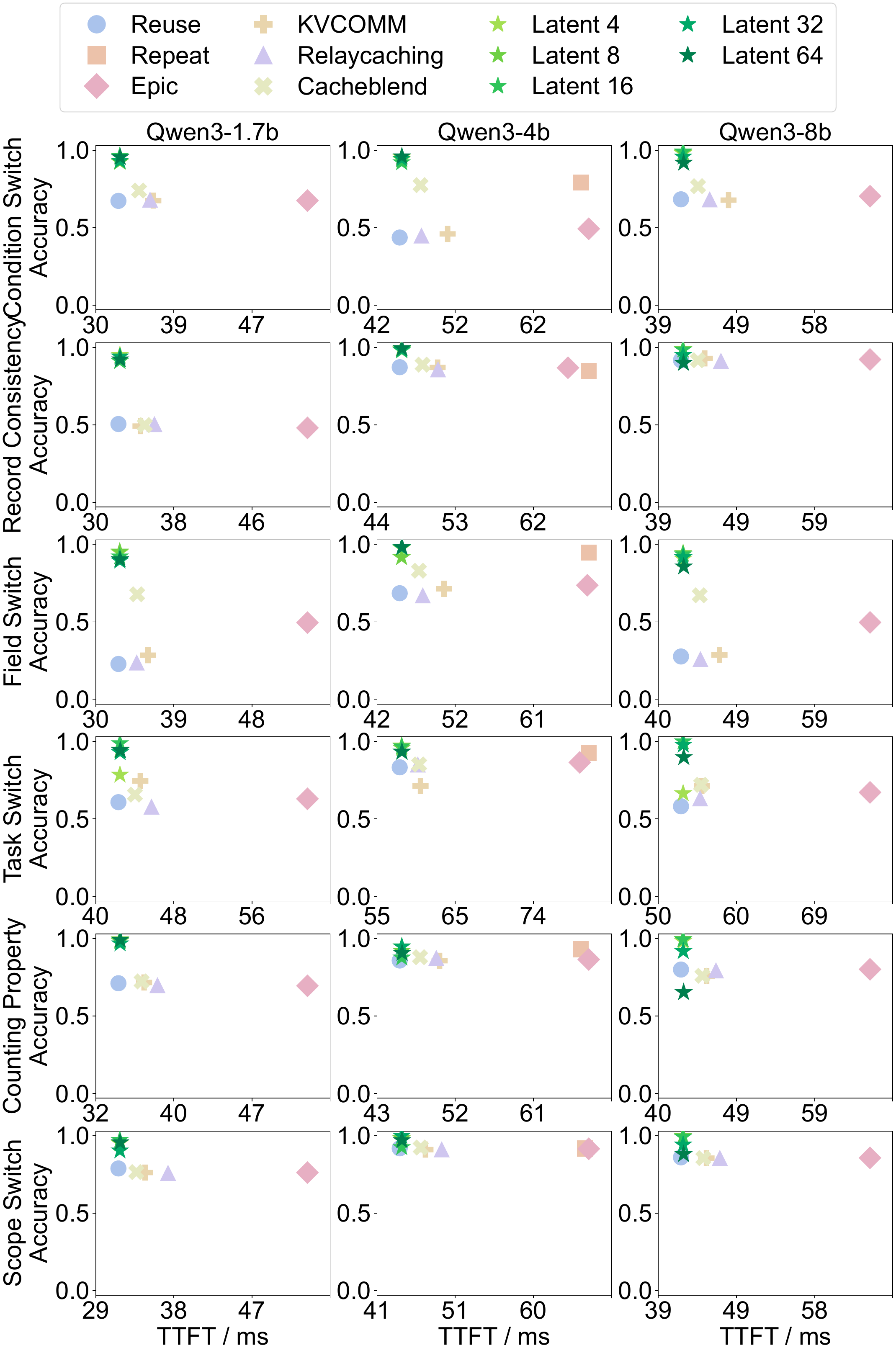}
    \caption{TTFT accuracy comparison across all tasks under reasoning settings using Qwen3-1.7B, 4B and 8B.}
    \label{fig:qwen3-4b-reasoning-ttft-acc-appendix-1}
\end{figure}

\begin{figure}[t]
    \centering
    \includegraphics[width=1\linewidth]{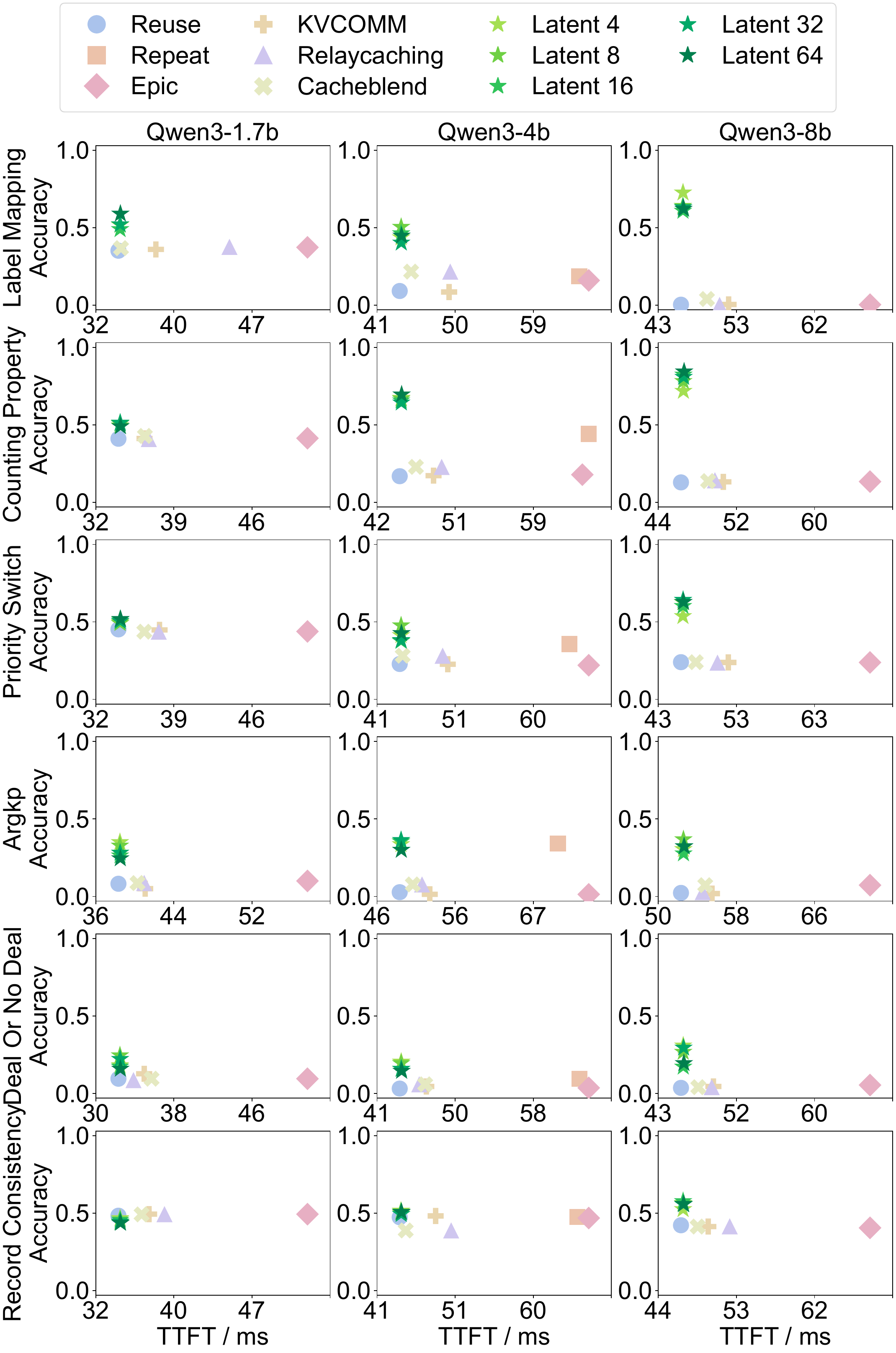}
    \caption{TTFT accuracy comparison across all tasks under no-reasoning settings using Qwen3-1.7B, 4B and 8B.}
    \label{fig:qwen3-4b-no-reasoning-ttft-acc-appendix-2}
\end{figure}

\begin{figure}[t]
    \centering
    \includegraphics[width=1\linewidth]{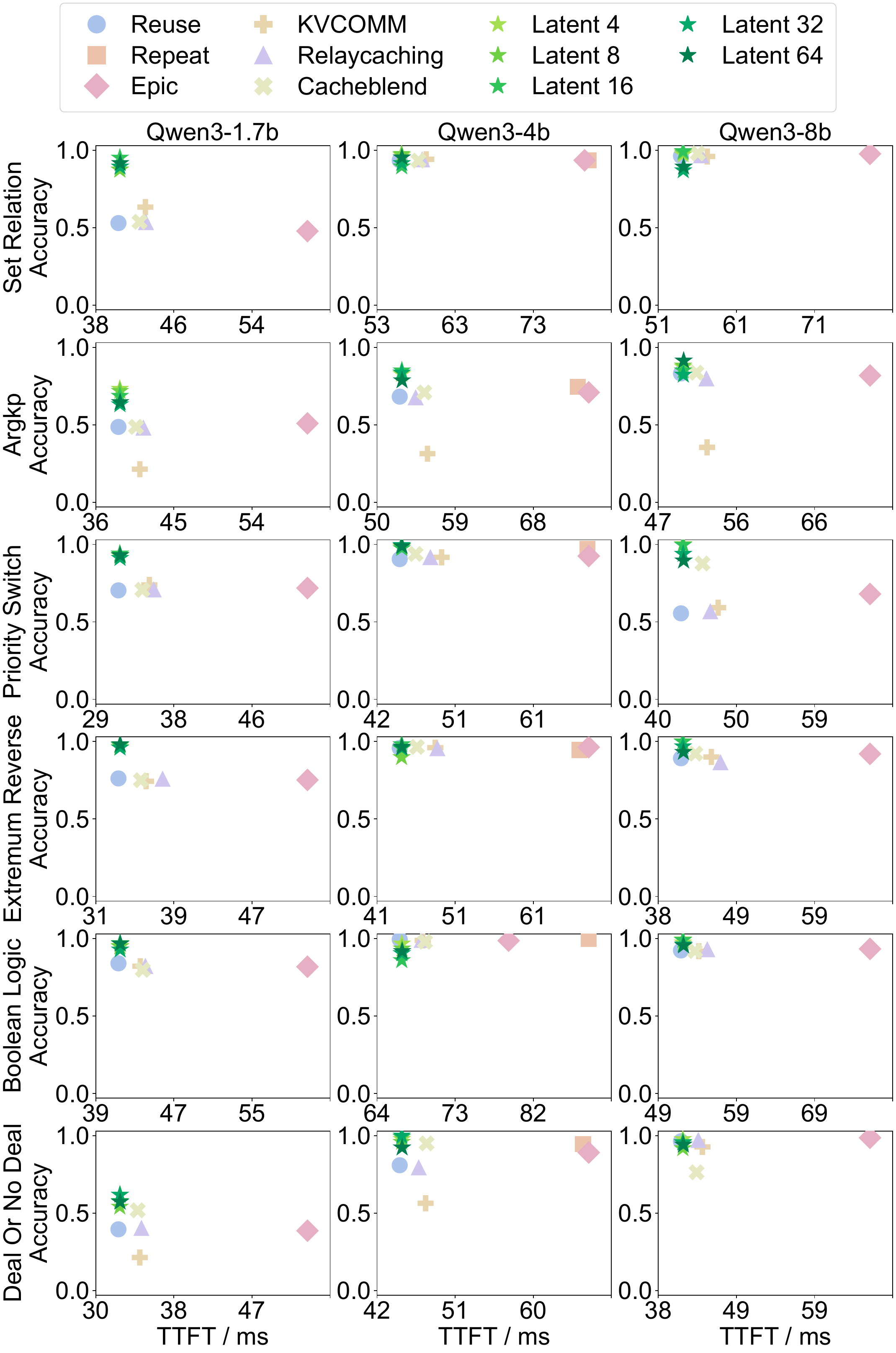}
    \caption{TTFT accuracy comparison across all tasks under reasoning settings using Qwen3-1.7B, 4B and 8B.}
    \label{fig:qwen3-4b-reasoning-ttft-acc-appendix-2}
\end{figure}

\begin{figure}[t]
    \centering
    \includegraphics[width=1\linewidth]{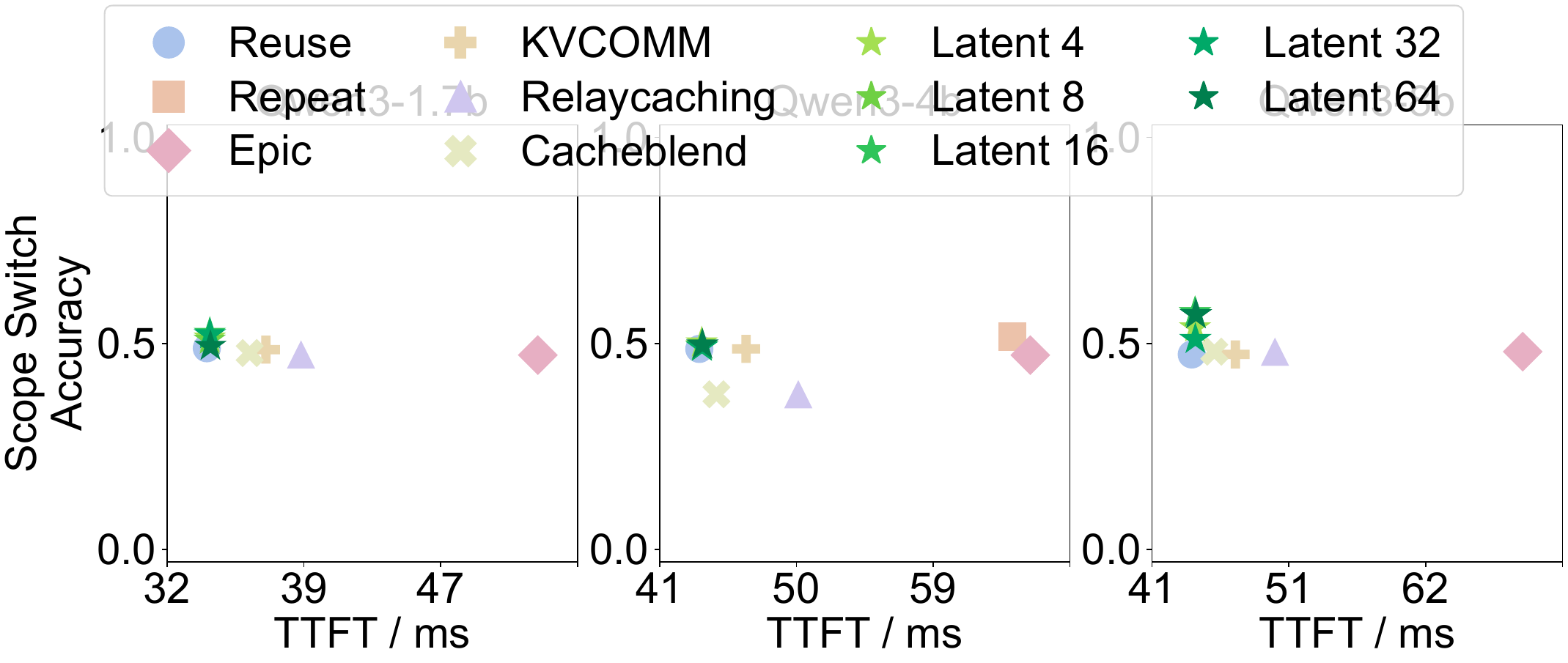}
    \caption{TTFT accuracy comparison across all tasks under no-reasoning settings using Qwen3-1.7B, 4B and 8B.}
    \label{fig:qwen3-4b-no-reasoning-ttft-acc-appendix-3}
\end{figure}

\begin{figure}[t]
    \centering
    \includegraphics[width=1\linewidth]{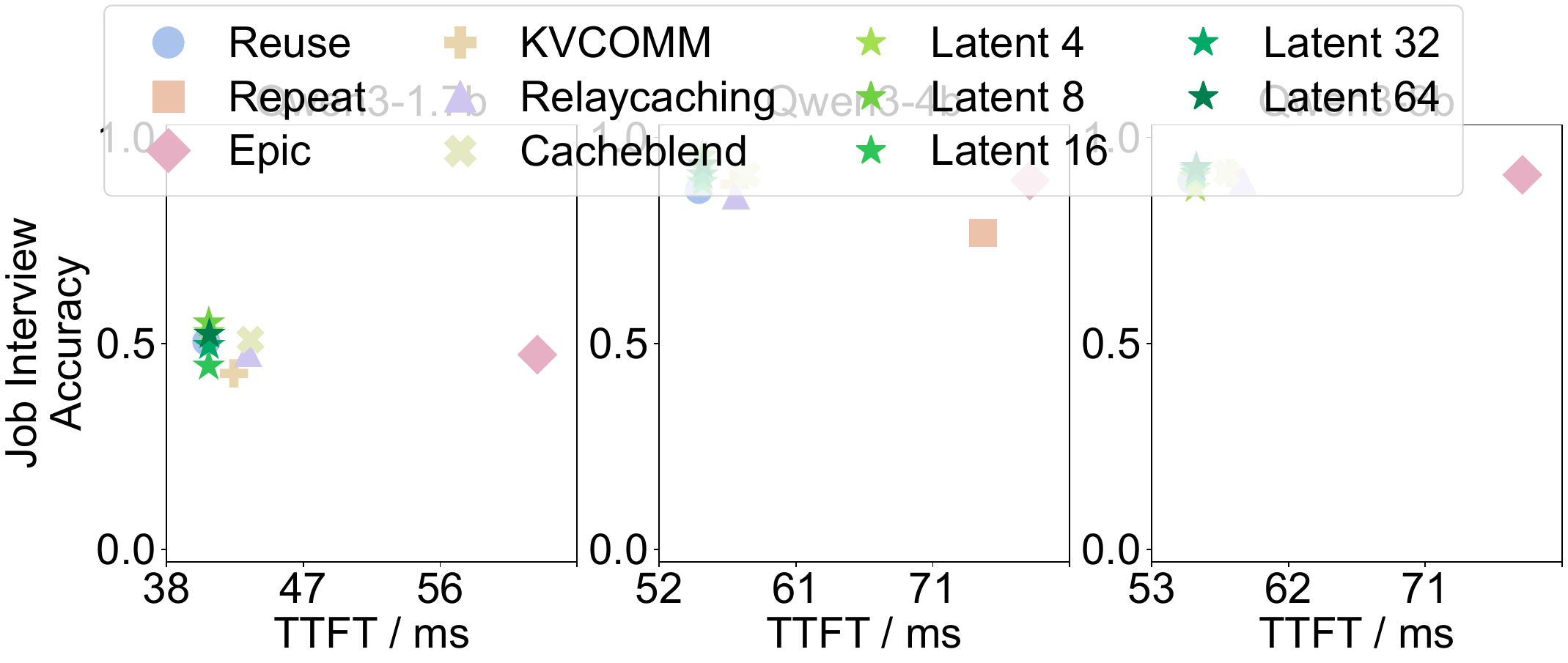}
    \caption{TTFT accuracy comparison across all tasks under reasoning settings using Qwen3-1.7B, 4B and 8B.}
    \label{fig:qwen3-4b-reasoning-ttft-acc-appendix-3}
\end{figure}

In this section we provide TTFT-accuracy figures for more tasks in Figure~\ref{fig:qwen3-4b-no-reasoning-ttft-acc-appendix-1}, ~\ref{fig:qwen3-4b-reasoning-ttft-acc-appendix-1},~\ref{fig:qwen3-4b-no-reasoning-ttft-acc-appendix-2}, ~\ref{fig:qwen3-4b-reasoning-ttft-acc-appendix-2}, ~\ref{fig:qwen3-4b-no-reasoning-ttft-acc-appendix-3} and ~\ref{fig:qwen3-4b-reasoning-ttft-acc-appendix-3}.

% 插入这些组图
% figure/task_ttft_accuracy/no_reasoning_appendix_1.pdf
% figure/task_ttft_accuracy/reasoning_appendix_1.pdf

% figure/task_ttft_accuracy/no_reasoning_appendix_2.pdf
% figure/task_ttft_accuracy/reasoning_appendix_2.pdf

% figure/task_ttft_accuracy/no_reasoning_appendix_3.pdf
% figure/task_ttft_accuracy/reasoning_appendix_3.pdf

\section{More Cases}
\label{appendix:case-study}
\begin{figure*}[!t]
    \centering
    \includegraphics[width=0.9\linewidth]{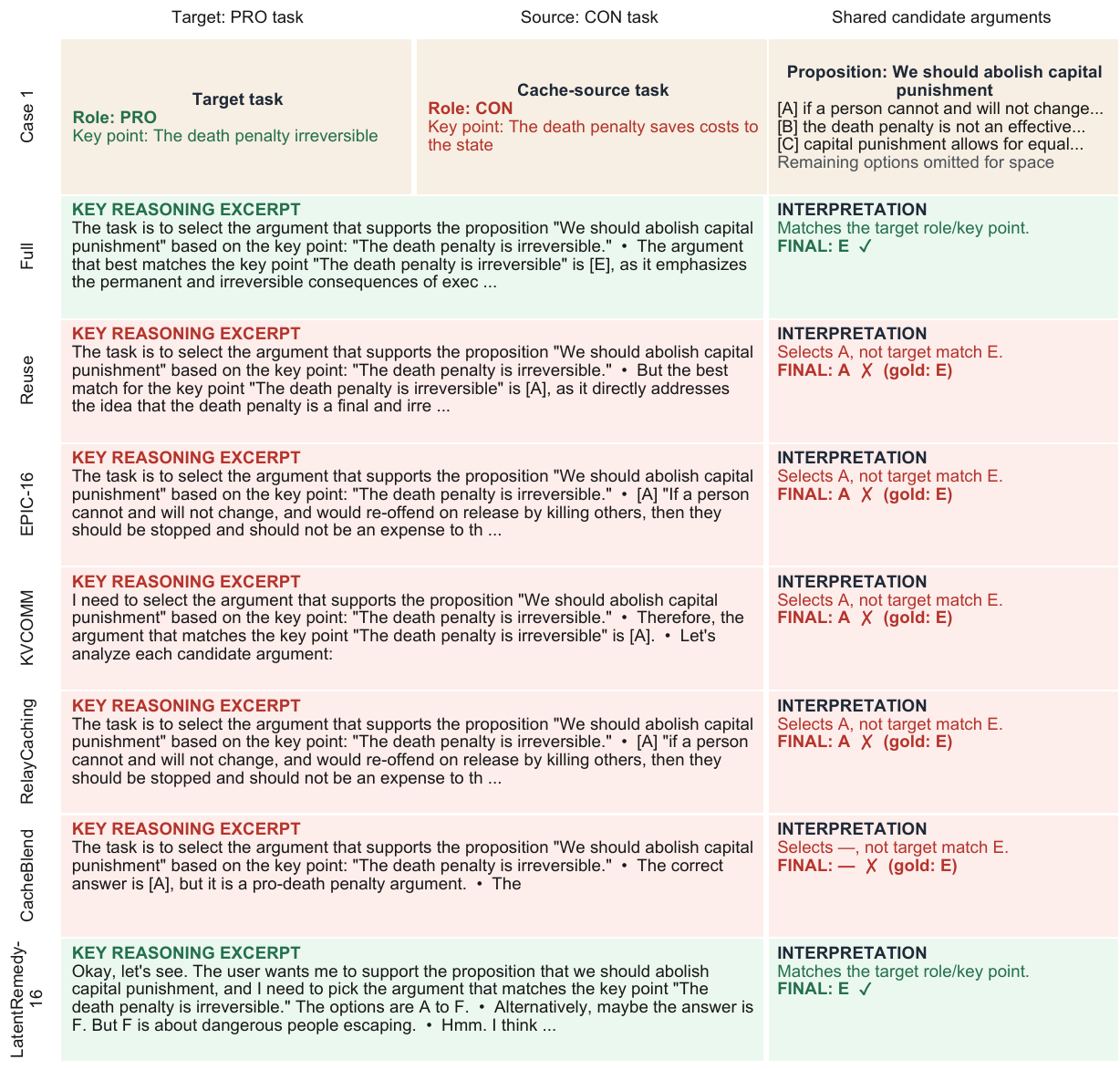}
    \caption{Case study of reasoning traces of different methods in the Argument Key Point task.}   
    \label{fig:case-study-argkp-1}
\end{figure*}

\begin{figure*}[!t]
    \centering
    \includegraphics[width=0.9\linewidth]{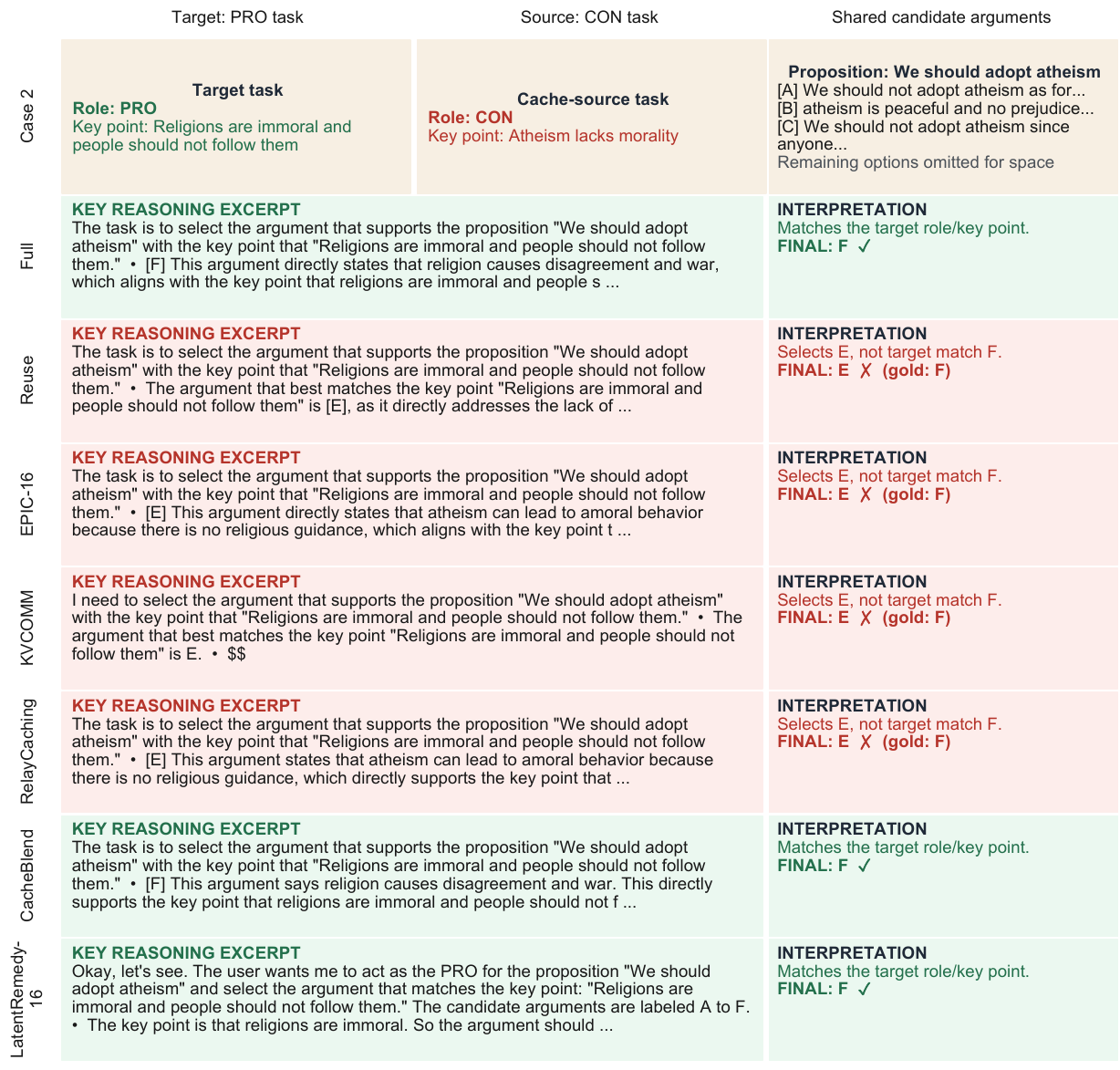}
    \caption{Case study of reasoning traces of different methods in the Argument Key Point task.}   
    \label{fig:case-study-argkp-2}
\end{figure*}

\begin{figure*}[!t]
    \centering
    \includegraphics[width=0.9\linewidth]{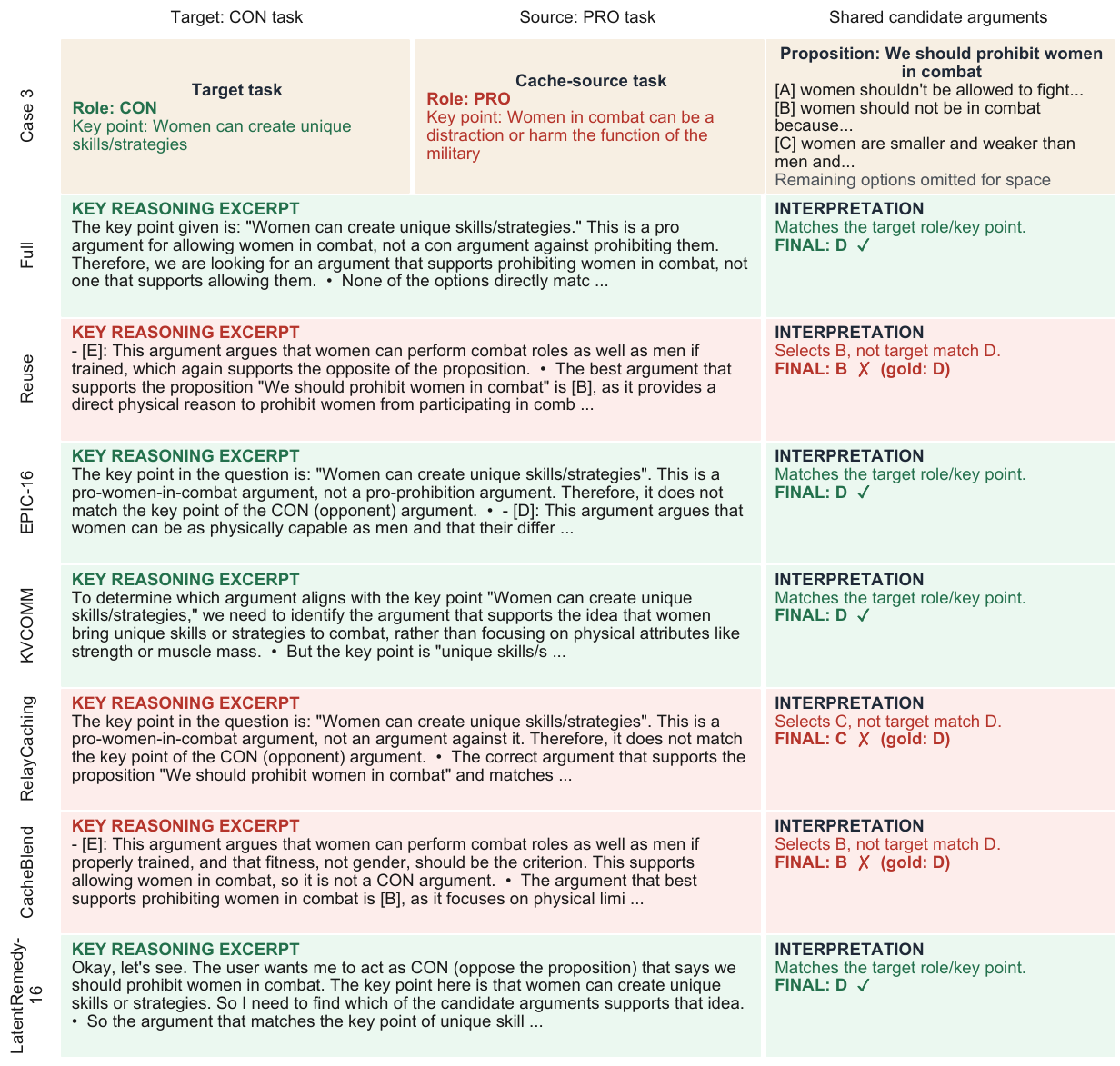}
    \caption{Case study of reasoning traces of different methods in the Argument Key Point task.}   
    \label{fig:case-study-argkp-3}
\end{figure*}

\begin{figure*}[!t]
    \centering
    \includegraphics[width=0.9\linewidth]{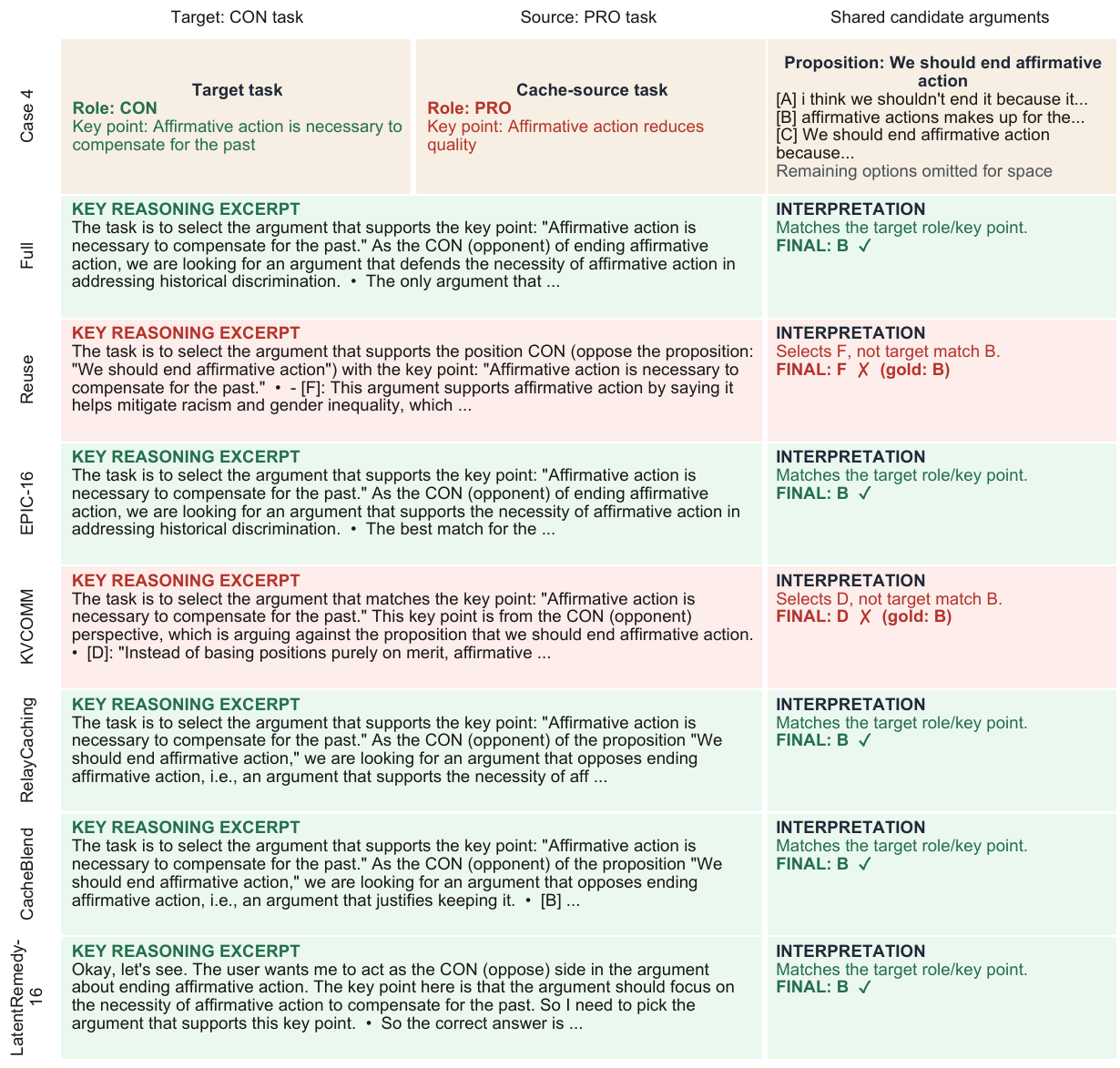}
    \caption{Case study of reasoning traces of different methods in the Argument Key Point task.}   
    \label{fig:case-study-argkp-4}
\end{figure*}

\begin{figure*}[!t]
    \centering
    \includegraphics[width=0.9\linewidth]{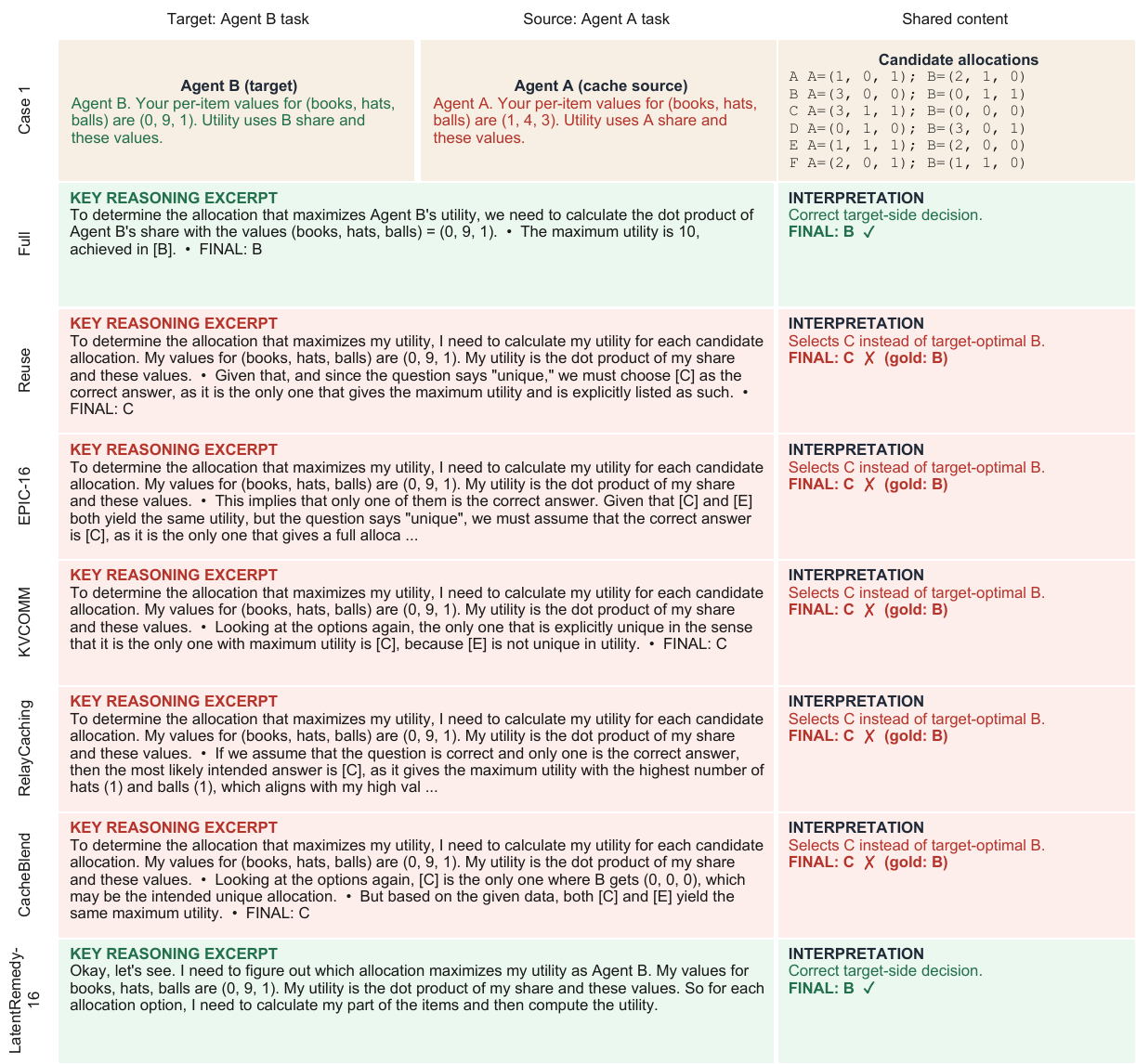}
    \caption{Case study of reasoning traces of different methods in the Deal or no deal task.}   
    \label{fig:case-study-deal-1}
\end{figure*}

\begin{figure*}[!t]
    \centering
    \includegraphics[width=0.9\linewidth]{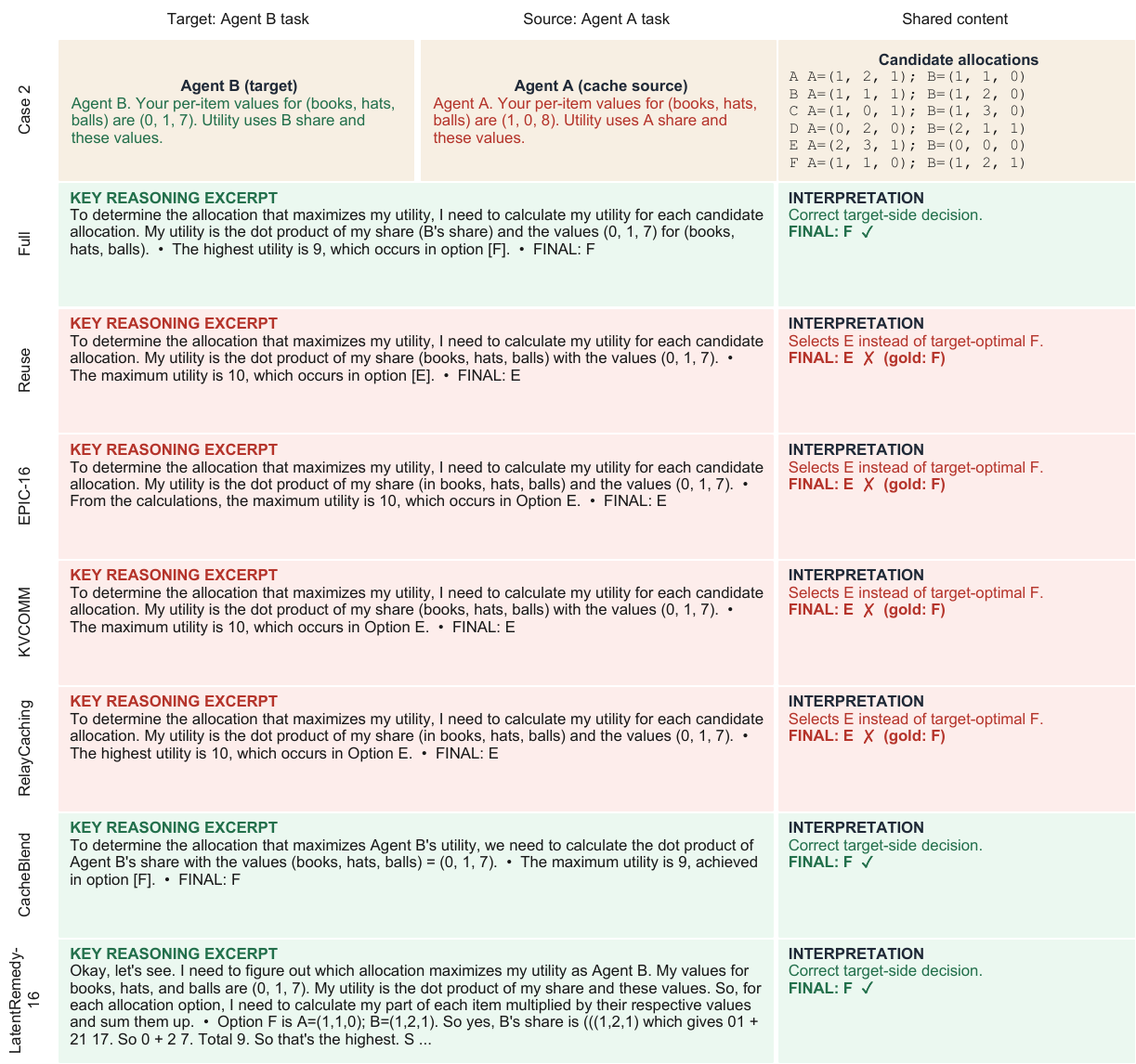}
    \caption{Case study of reasoning traces of different methods in the Deal or no deal task.}   
    \label{fig:case-study-deal-2}
\end{figure*}

\begin{figure*}[!t]
    \centering
    \includegraphics[width=0.9\linewidth]{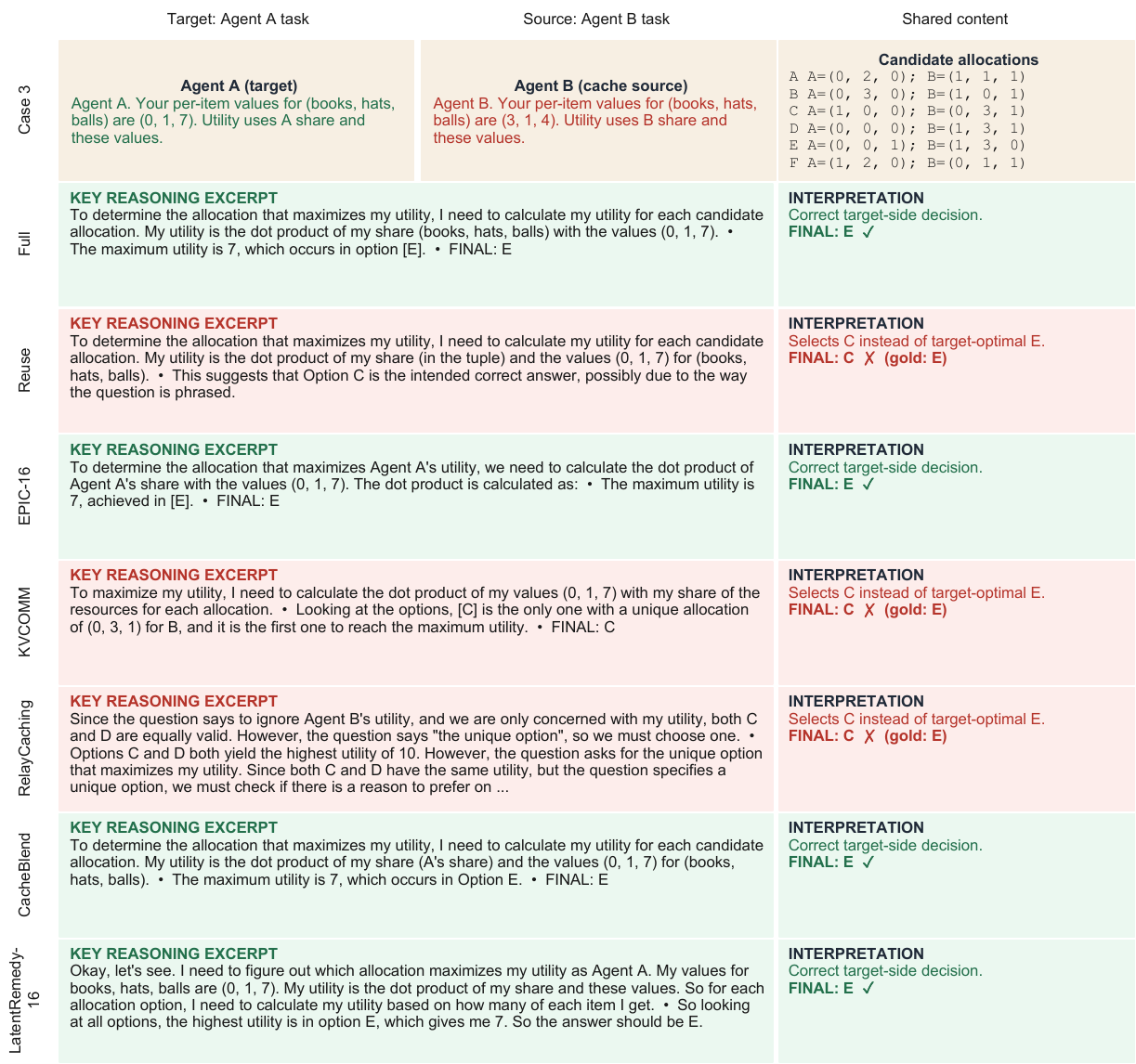}
    \caption{Case study of reasoning traces of different methods in the Deal or no deal task.}   
    \label{fig:case-study-deal-3}
\end{figure*}

\begin{figure*}[!t]
    \centering
    \includegraphics[width=0.9\linewidth]{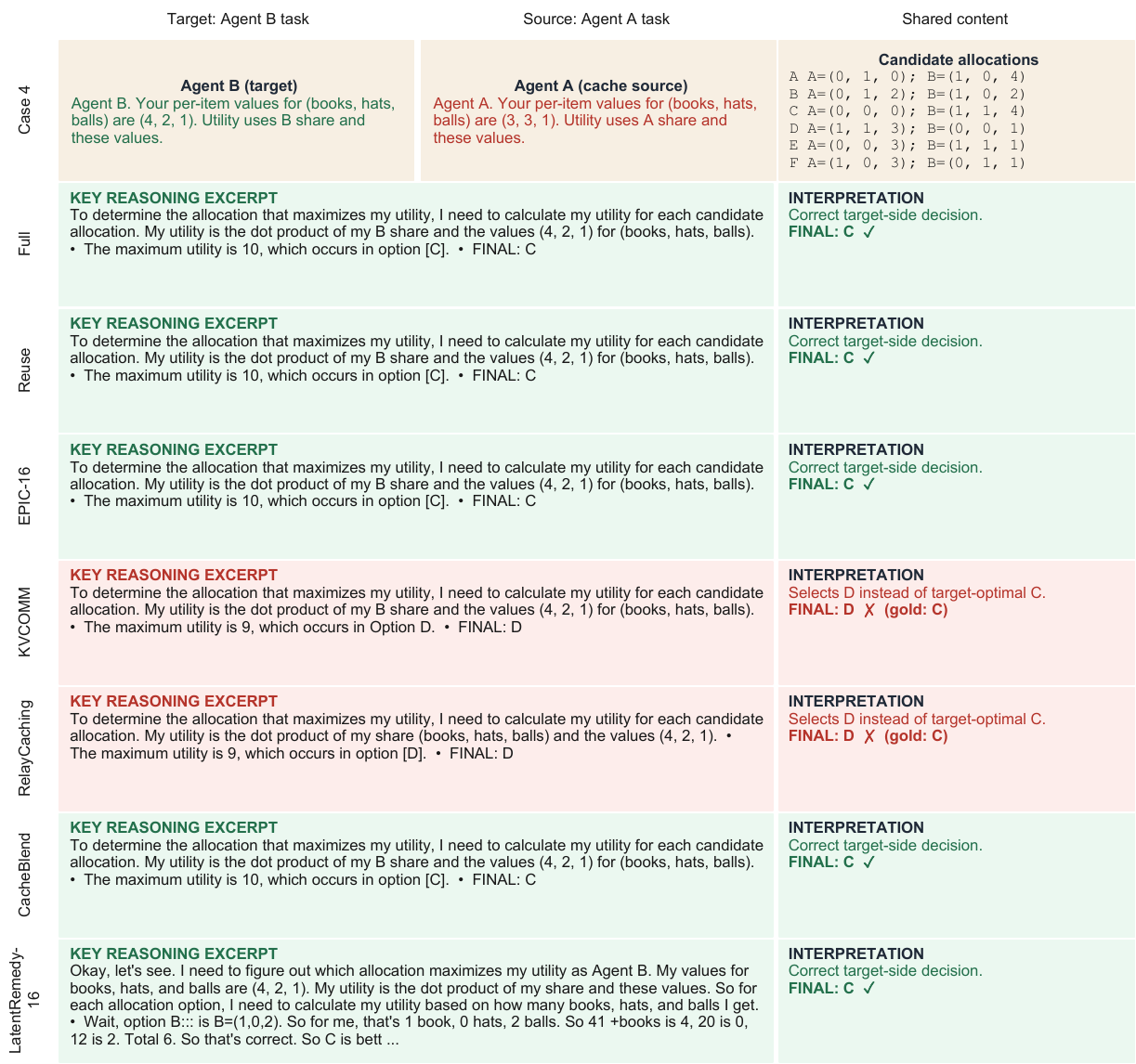}
    \caption{Case study of reasoning traces of different methods in the Deal or no deal task.}   
    \label{fig:case-study-deal-4}
\end{figure*}

In this section we provide more cases regarding different methods and tasks in Figure~\ref{fig:case-study-argkp-1},~\ref{fig:case-study-argkp-2},~\ref{fig:case-study-argkp-3},~\ref{fig:case-study-argkp-4},~\ref{fig:case-study-deal-1},~\ref{fig:case-study-deal-2},~\ref{fig:case-study-deal-3},~\ref{fig:case-study-deal-4}. 

\section{Dataset Overview}
\label{appendix:dataset-overview}

\noindent
\textbf{Design principles.}
% 这里主要讲我们设计的数据集目的是拥有两个任务，共同的文本块；以便让agent A和B读取和理解相同的文本块，并在此基础上进行推理和决策。我们希望数据集能够覆盖不同的任务类型和难度，以便评估模型在不同场景下的表现。
As current KV cache reuse methods are evaluated mainly on cooperative tasks such as solving problems in MMLU, HumanEval, with RAG multi-turn discussion, we aim to develop a dataset that can be evaluated in multi-agent systems where agents are dispatched with different tasks with the same context provided. To this end, we introduce ConflictInstruct, a benchmark that evaluates agent performance when multi-agent systems are reusing the same context KV cache but with different goals. The dataset consists of various task families that can be dispatched in conflicted ways such as selection criteria, data grouping and logical induction. Specifically, agent A and agent B are given different instructions $p_A$ and $p_B$, continued with the same shared context $s$, and then required to produce different outputs $y_A$ and $y_B$. The dataset together with code is available at https://anonymous.4open.science/r/kvreuse-F7E6/.

\noindent
\textbf{Generation process.}
% 这里主要讲我们如何采用evolinstruct生成数据集。主要介绍我们的要求是什么、生成的数据的验证方式是什么。
The dataset is mainly driven by EvolInstruct with seed tasks manually created, then expanded by LLMs~\cite{zeng-etal-2024-automatic}. Iteratively, we prompt LLM to propose either new operators on existing data forms, new scenarios in the same task form, or tasks and data that are different from existing examples. After generation, all tasks are going through full-prefill evaluation to make sure the correctness of the data with best effort. We discard those with inappropriate formats and incorrect answers, resulting in finally 240 task families with 30 pairs of instances per family.

\noindent
\textbf{Dataset statistics.}
% 这里主要讲我们数据集的统计信息，列一个表，包括任务名字和对应的任务指令长度、共享文本长度。
The balanced corpus contains 240 task families with 30 paired instances per family, giving 7,200 pairs in total. Table~\ref{tab:conflictinstruct-category-statistics} summarizes the families in terms of general task types and token length distribution in terms of task instruction and shared context.  Overall, the ConflictInstruct data has a mean instruction length of 88.3 tokens and a mean shared-block length of 260.6 tokens, with ranges of 19--254 and 5--2,200 tokens, respectively. Task variety spreads across audited algorithmic families, basic instruction switches, scene-augmented switches, structured reasoning, long-form role conflicts, argument-perspective scenes, allocation-utility scenes, and offer-evaluation scenes. 

\begin{table*}[t]
    \centering
    \small
    \setlength{\tabcolsep}{5pt}
    \begin{tabular}{@{}lrrrrr@{}}
        \toprule
        Task group & Families & Train / Test & Pairs & Instruction & Shared block \\
        \midrule
        Audited algorithmic families & 95 & 69 / 26 & 2,850 & 123.7 & 117.7 \\
        Basic instruction switches & 16 & 13 / 3 & 480 & 46.7 & 170.0 \\
        Scene-augmented switches & 40 & 30 / 10 & 1,200 & 53.1 & 167.9 \\
        Structured reasoning & 68 & 52 / 16 & 2,040 & 85.5 & 439.5 \\
        Long-form role conflicts & 6 & 4 / 2 & 180 & 33.6 & 1,775.7 \\
        Argument-perspective scenes & 5 & 5 / 0 & 150 & 20.1 & 84.2 \\
        Allocation-utility scenes & 5 & 5 / 0 & 150 & 48.0 & 115.0 \\
        Offer-evaluation scenes & 5 & 2 / 3 & 150 & 41.1 & 77.6 \\
        \midrule
        Total / weighted mean & 240 & 180 / 60 & 7,200 & 88.3 & 260.6 \\
        \bottomrule
    \end{tabular}
    \caption{Statistics of ConflictInstruct across different task groups.}
    \label{tab:conflictinstruct-category-statistics}
\end{table*}

\noindent
\textbf{Dataset splits.}
% 我们按照任务family来划分75%训练，25%测试。
We split the data by task family, with 25\% families for testing, and 75\% for development. Each family constains 30 samples, and no overlap exists between the two splits in terms of family. Optionally, ArgKP, Deal or No Deal, and Job Interview can be used as real-world OOD task evaluation.

We partition the corpus at the task-family level using a fixed random seed (20261005): 180 families (75\%) form train75 and the remaining 60 families (25\%) form test25. For each family, we retain the first 30 instances in the frozen source ordering, yielding 5,400 training pairs and 1,800 unseen task test pairs (Table~\ref{tab:conflictinstruct-split-statistics}). 

\begin{table}[t]
    \centering
    \small
    \setlength{\tabcolsep}{4pt}
    \begin{tabular}{@{}lrrrr@{}}
        \toprule
        Split & Families & Pairs & Instruction & Shared \\
        \midrule
        train75 & 180 & 5,400 & 86.7 & 253.7 \\
        test25 & 60 & 1,800 & 93.1 & 281.0 \\
        \bottomrule
    \end{tabular}
    \caption{Family-level 75/25 split of ConflictInstruct. Instruction and Shared are mean token lengths under the measurement convention in Table~\ref{tab:conflictinstruct-category-statistics}.}
    \label{tab:conflictinstruct-split-statistics}
\end{table}

\end{document}